\documentclass[11pt]{article}
\usepackage[margin=1in]{geometry}
\usepackage{times}
\usepackage{natbib}
\usepackage{amsmath,amsfonts,bm}

\def\eqref#1{equation~\ref{#1}}
\def\1{\bm{1}}

\DeclareMathAlphabet{\mathsfit}{\encodingdefault}{\sfdefault}{m}{sl}
\SetMathAlphabet{\mathsfit}{bold}{\encodingdefault}{\sfdefault}{bx}{n}

\usepackage{graphicx}
\usepackage{float}
\usepackage{wrapfig}

\usepackage{booktabs}
\usepackage{longtable}
\usepackage{multirow}
\usepackage{titletoc}
\usepackage{tabularx}
\usepackage[most]{tcolorbox}
\usepackage{hyperref}
\usepackage{url}

\graphicspath{{figures/}}

\title{Trace Evolution Bench: Measuring Whether Agents Improve\\
from Their Own Environment Traces}

\author{Fangfei Li,$^{1}$ Chenyang Zhao,$^{1}$ Long Wang,$^{1}$ Feng Tian,$^{1}$ Lv Guo$^{1,*}$\\
$^{1}$MatrixOrigin, China\\
$^{*}$Corresponding author.\\
\texttt{lifangfei@matrixorigin.cn}, \texttt{zhaochenyang@matrixorigin.cn},\\
\texttt{wanglong@matrixorigin.io}, \texttt{tianfeng@matrixorigin.cn},\\
\texttt{guolv@matrixorigin.cn}$^{*}$}
\date{}

\begin{document}

\maketitle

\begin{abstract}
Self-evolving agents are often judged by endpoint Gain (final held-out success minus round-0 success), which can conflate true evolution with frozen rescoring variance. We present Trace Evolution Bench (TEB), an evaluation protocol that isolates attributable improvement through four controls: locked held-out identifiers, validation-gated accept/rollback, repeated frozen \texttt{no\_evolve} rescoring, and seed-matched paired excess against each seed's own noise band. TEB includes three frozen 80/20/40 packs, with coding as the primary scoreboard because success is determined by hidden unit tests in a closed workspace. On coding (40 held-out tasks), DeepSeek v4 Pro under \texttt{no\_evolve} keeps version $M_0$ yet shifts by 7.5, 5.0, and 12.5 percentage points across seeds, showing that endpoint Gain alone overstates progress. Under this control, seven of nine DeepSeek text-update cells exceed their seed bands, but the pattern does not reproduce on GLM and MiniMax; no method remains above its own band on every seed of all three models. On separate packs, office DeepSeek does not exceed its seed bands (widest 17.5 points), and manufacturing has no paired positive excess beyond its seed band. These results position TEB as a reproducible attribution contract for self-evolving agents.
\end{abstract}

\newcommand{\tebarxiv}{}
\section{Introduction}

Most agent benchmarks freeze a policy and score static capability~\citep{liu2024agentbench,jimenez2024swebench,yao2023react,zhou2024webarena,mialon2024gaia,shridhar2021alfworld}. Self-evolving systems pose a different measurement problem: whether post-update improvement is attributable to the evolution procedure rather than to evaluation noise or protocol mismatch~\citep{silver2017alphago}. In practice, attribution is confounded when teacher traces substitute for self-generated traces~\citep{ross2011dagger}, when evaluation targets a checkpoint different from the artifact returned by \texttt{evolve()}, when held-out ids leak into the update~\citep{magar2022contamination}, or when text updates and weight updates are compared under one budget axis~\citep{henderson2018matters}. Reporting Success~\% alone can hide weak round-0 baselines~\citep{liang2023helm}; reporting Gain from one stochastic round-0 pass can credit frozen rescoring variance as evolution~\citep{dodge2019show}. These issues motivate an attribution-first contract, not only a final-score report.

Trace Evolution Bench (TEB) operationalizes this contract with locked held-out ids, validation-gated accept/rollback, frozen \texttt{no\_evolve} rescoring, and seed-matched paired excess. We report coding as the primary scoreboard because success is determined by hidden unit tests in a closed workspace: tests, rather than phrasing, decide each pass. This setting is harder to game than office's rule-bounded delivery and broader than manufacturing table QA. The three scored methods were fixed before held-out reading, via an office validation screen whose mean gains were not positive (Section~\ref{sec:coding}). On coding, DeepSeek v4 Pro~\citep{deepseek2026v4} \texttt{no\_evolve} keeps version $M_0$ yet moves $7.5$, $5.0$, and $12.5$ percentage points across four passes of the same 40 held-out tasks (Table~\ref{tab:coding}), showing that endpoint Gain still contains frozen rescore. Under paired excess, seven of nine DeepSeek text-update cells lie above their own seed bands, while \texttt{gepa\_llm} seed 44 moves down by the width of its band (Table~\ref{tab:codingexcess}, Figure~\ref{fig:codingexcess}). GLM and MiniMax do not reproduce that seven-of-nine pattern (Figure~\ref{fig:codingexcessother}, Table~\ref{tab:codingstatus}); no method stays above its own band on every seed of all three models. GLM has two paired cells above their own bands, and MiniMax \texttt{expel\_llm} seed 44 is a third, with final regression $0.30$. Office and manufacturing are interpreted as separate control packs on their own bands: office DeepSeek does not exceed its bands (widest $17.5$ pp; Section~\ref{sec:officectr}), and manufacturing has no paired positive excess beyond its seed band (Section~\ref{sec:mfgbody}).

\paragraph{Contributions.}
(1)~An attribution-first evaluation contract for self-evolving agents: locked held-out ids, validation-gated accept/rollback, frozen \texttt{no\_evolve} rescoring, and separate budget classes for B0 text updates versus B2 weight updates.
(2)~A reproducible diagnostic layer for gain claims: per-seed frozen noise bands, seed-matched paired excess, round-0 pairing checks, and explicit regression/eligibility markers that separate scoring evidence from ranking policy.
(3)~Cross-model evidence on the coding primary scoreboard: DeepSeek shows seven of nine cells above their own bands, but this pattern does not reproduce on GLM and MiniMax; no method stays above its own band on every seed of all three models.
(4)~Three frozen 80/20/40 packs with separate interpretation lanes: coding as primary evidence, office and manufacturing as independent controls on their own seed bands.

\section{Related work}

ReAct-style loops~\citep{yao2023react} and agent suites~\citep{liu2024agentbench,jimenez2024swebench} grade a frozen policy. Reflexion~\citep{shinn2023reflexion}, ExpeL~\citep{zhao2024expel}, self-refinement~\citep{madaan2023selfrefine}, workflow memory~\citep{wang2024awm}, prompt evolution~\citep{agrawal2025gepa}, skill libraries~\citep{wang2023voyager}, and open-ended mutation~\citep{zhang2025dgm} are systems TEB can host. The adapters are method-inspired text updates on one frozen actor, and the same three methods are scored on coding, office, and manufacturing. The quantity compared with those systems is whether an update beats a frozen rescore, not their published scores.

Three protocol choices separate TEB from that set. The held-out ids never enter \texttt{evolve()}. Neighbors that mix the test ids into the update, or that do not say which split is locked, are not the same contract~\citep{magar2022contamination}. A validation gate accepts a proposal only when validation success does not fall. Held-out success is scored after that decision, on the accepted artifact, and is not the accept test. Weight updates are a separate budget class. A B0 text update is not ranked against a B2 weight update, including LoRA-style training~\citep{hu2022lora}.

Table~\ref{tab:evalq} is the evaluation question.

\begin{table}[ht]
\caption{Evaluation target of each benchmark. A frozen rescore band is the range of repeated held-out passes of one unchanged policy.}
\label{tab:evalq}
\begin{center}
\scriptsize
\setlength{\tabcolsep}{3pt}
\begin{tabular}{@{}p{0.28\linewidth}p{0.32\linewidth}p{0.32\linewidth}@{}}
\toprule
Benchmark & Scored unit & Frozen rescore band \\
\midrule
SWE-bench, AgentBench~\citep{jimenez2024swebench,liu2024agentbench} & One episode & No \\
LifelongAgentBench~\citep{zheng2025lifelongagentbench} & A task sequence & No \\
SEA-Eval~\citep{jiang2026seaeval} & A task sequence & No \\
EvoAgentBench~\citep{gao2026evoagentbench} & Transfer on an ability-supported split & No; one vanilla condition \\
EvoPathBench~\citep{lin2026evopath} & Capability checkpoints over ordered streams & No \\
S$^3$Gym~\citep{shi2026s3gym} & Strict held-out evaluation & No \\
TEB & The same locked held-out ids & Yes; four \texttt{no\_evolve} passes \\
\bottomrule
\end{tabular}
\end{center}
\end{table}

Table~\ref{tab:neighbors} is the feature contrast under binary coding, using a stated-evidence review. \checkmark\ marks explicit support in the paper's main protocol, and \texttimes\ marks all other cases (including not stated). \citet{nguyen2026rsea} commit a candidate only when held-out performance does not regress, so the validation-gate and held-out-lock cells are crosses. \citet{zheng2026seagym} separate training, validation, and held-out views but do not state which split accepts an update. \citet{shi2026s3gym} separate exploration from strict held-out evaluation and compare history, summary-memory, and training pathways in one framework. The freeze-weights row is a cross where the benchmark trains: \citet{meng2026rsibench}, \citet{kang2026trace}, \citet{yang2026openmle}, \citet{sebench2026}, \citet{rank2026posttrainbench}, and \citet{shi2026s3gym}. PostTrainBench also forbids training on the test set and scores the submitted checkpoint. Those papers also discuss transfer, business tasks, frozen artifacts, retained experience, and whether one financial case helps the next; none of those properties is a row of Table~\ref{tab:neighbors}~\citep{gao2026evoagentbench,zhou2026gdpevo,lin2026evopath,pastbench2026,finevo2026}. \citet{zheng2025lifelongagentbench} and \citet{jiang2026seaeval} evaluate agents over task sequences rather than single episodes; \citet{liu2026skillfeedback} studies skill feedback over rounds, and \citet{shang2026backfires} gates skill updates before commit.

\begin{table}[ht]
\caption{Protocol mechanics contrast (binary coding). \checkmark\ means the property is explicitly supported in the paper's main protocol; \texttimes\ means otherwise (including not stated). In ``Carrier flex. in this paper (m/p/c/w)'', m/p/c/w = memory/prompt/code/weight. Abbreviations: SEA (SEAGym), Evo (EvoAgentBench), GDP (GDPevo), RSI (RSIBench-Data), PTB (PostTrainBench), TR (TRACE), MLE (OpenMLE), EPB (EvoPathBench), S3 (S$^3$Gym), SEB (SE-Bench), PAST (PAST-Bench), Fin (FinEvo-Bench).}
\label{tab:neighbors}
\begin{center}
\scriptsize
\setlength{\tabcolsep}{2pt}
\renewcommand{\arraystretch}{1.03}
\begin{tabular}{@{}p{0.17\linewidth}cccccccccccccc@{}}
\toprule
Property & TEB & RSEA & SEA & Evo & GDP & RSI & PTB & TR & MLE & EPB & S3 & SEB & PAST & Fin \\
\midrule
Own traces & \checkmark & \checkmark & \checkmark & \checkmark & \checkmark & \texttimes & \checkmark & \checkmark & \checkmark & \checkmark & \checkmark & \texttimes & \checkmark & \checkmark \\
Multi-round & \checkmark & \checkmark & \checkmark & \texttimes & \texttimes & \checkmark & \texttimes & \texttimes & \texttimes & \checkmark & \checkmark & \texttimes & \checkmark & \checkmark \\
Carrier flex. in this paper (m/p/c/w) & \texttimes & \texttimes & \texttimes & \texttimes & \texttimes & \texttimes & \texttimes & \texttimes & \texttimes & \texttimes & \texttimes & \texttimes & \texttimes & \texttimes \\
Explore/eval split & \checkmark & \texttimes & \texttimes & \checkmark & \texttimes & \texttimes & \checkmark & \texttimes & \texttimes & \checkmark & \checkmark & \texttimes & \texttimes & \texttimes \\
Validation gate & \checkmark & \texttimes & \texttimes & \texttimes & \texttimes & \texttimes & \texttimes & \texttimes & \texttimes & \texttimes & \texttimes & \texttimes & \texttimes & \texttimes \\
Held-out lock & \checkmark & \texttimes & \texttimes & \checkmark & \texttimes & \texttimes & \checkmark & \texttimes & \texttimes & \checkmark & \checkmark & \texttimes & \texttimes & \texttimes \\
Weight freeze & \checkmark & \checkmark & \checkmark & \checkmark & \texttimes & \texttimes & \texttimes & \texttimes & \texttimes & \checkmark & \texttimes & \texttimes & \checkmark & \checkmark \\
Weight slot & \checkmark & \texttimes & \texttimes & \texttimes & \texttimes & \texttimes & \texttimes & \texttimes & \texttimes & \texttimes & \texttimes & \texttimes & \texttimes & \texttimes \\
\bottomrule
\end{tabular}
\renewcommand{\arraystretch}{1.0}
\end{center}
\end{table}

\section{Protocol}
\label{sec:protocol}

Figure~\ref{fig:pipeline} is the measurement. Figure~\ref{fig:protocol} is one round of it. Round $0$ scores $M_0$ on held-out and freezes Gain's baseline. Each later round collects a disjoint slice of evolution episodes, calls \texttt{evolve()} once, accepts the proposal iff validation success is at least the current agent's (\texttt{min\_delta}${=}0$), otherwise rolls back, then scores the accepted agent on the same held-out ids. Held-out success never enters \texttt{evolve()}.

\ifdefined\tebarxiv
\begin{figure}[t]
\else
\begin{figure}[H]
\fi
\centering
\includegraphics[width=\linewidth]{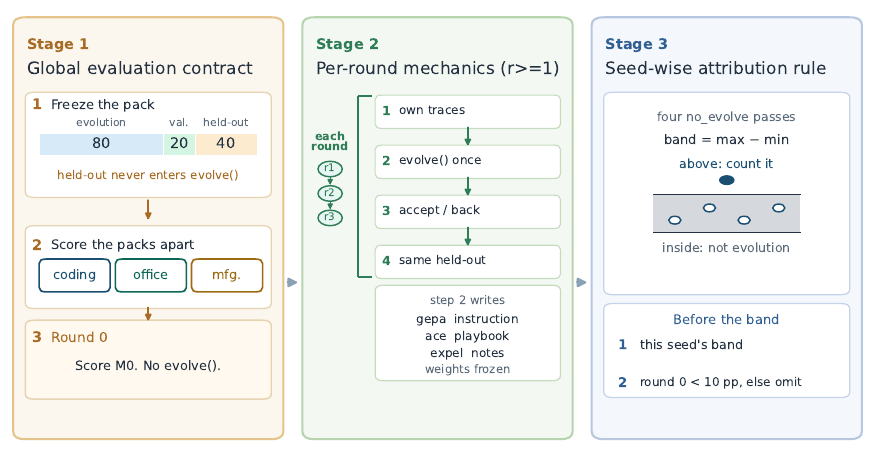}
\caption{TEB global evaluation contract across rounds. Round 0 scores \(M_0\) and does not update. Rounds 1--3 each accept or roll back one text update, then rescore the same held-out. Excess is read against that seed's own four-pass band.}
\label{fig:pipeline}
\end{figure}

Let $\mathrm{Succ}(M)=\#\{\text{held-out pass}\}/n$ be a rate on the locked held-out ids. Percentage points are $100$ times a rate difference:
\begin{equation}
\mathrm{Gain}(M_r)=100\times\bigl(\mathrm{Succ}(M_r)-\mathrm{Succ}(M_0)\bigr),
\label{eq:gain}
\end{equation}
with this-run $M_0$. In these equations $M_0$ is that round-0 score. A version label $M_0, M_1, \ldots$ names the accepted artifact, and \texttt{no\_evolve} keeps version $M_0$ while the score still moves. A proposal is accepted when validation success is at least the current agent's ($\mathrm{min\_delta}=0$); otherwise it is rolled back. Held-out ids do not enter \texttt{evolve()}. For actor $a$, pack $e$, and seed $s$, the noise band is the range of the four frozen \texttt{no\_evolve} held-out rates $k\in\{0,1,2,3\}$:
\begin{equation}
B(a,e,s)=100\times\Bigl(\max_k \mathrm{Succ}(M^{(0)}_{a,e,s,k})-\min_k \mathrm{Succ}(M^{(0)}_{a,e,s,k})\Bigr).
\label{eq:band}
\end{equation}
Paired excess uses the same actor, pack, and seed:
\begin{equation}
\mathrm{Excess}=\mathrm{Gain}_{\mathrm{method}}(a,e,s)-\mathrm{Gain}_{\texttt{no\_evolve}}(a,e,s).
\label{eq:excess}
\end{equation}
An excess inside that seed's own $B(a,e,s)$ is not reported as evolution. A method whose round-0 held-out rate differs from the same-seed frozen round 0 by 10 pp or more is not read as the same starting draw. Each seed is compared with its own $B(a,e,s)$. Office DeepSeek's widest seed band is $17.5$ pp, and seed 43's own band is $7.5$ pp. A value inside the widest band can still sit on seed 43's band. Manufacturing keeps its own bands. Regression is, among tasks passed on the previous held-out pass, the fraction that fail on the final pass. A rate above $0.20$ is stored on the cell. The excess comparison keeps the cell either way. Coding is the primary scoreboard. The evolution-token cap is $500{,}000$ per round, summed from per-step usage on evolution tasks only. Validation, held-out, and calls inside \texttt{evolve()} are not added. Crossing the cap skips \texttt{evolve()} for that round, marks \texttt{over\_budget}, and skips later rounds when the violation is recorded before the next round.

\ifdefined\tebarxiv
\begin{figure}[t]
\else
\begin{figure}[H]
\fi
\centering
\includegraphics[width=0.90\linewidth]{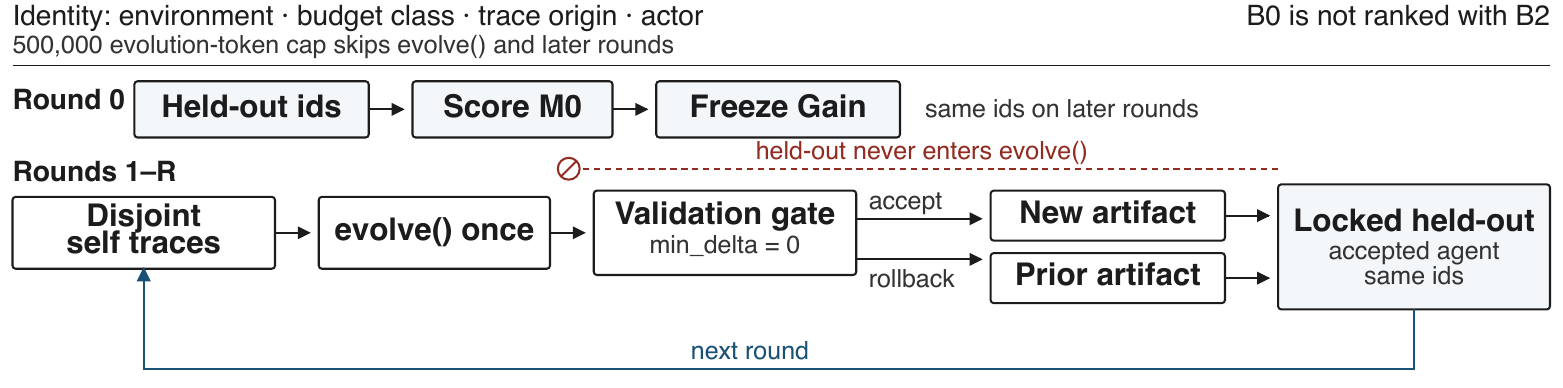}
\caption{Per-round accept/rollback mechanics (this loop is rounds \(r \ge 1\) in Figure~\ref{fig:pipeline}). Round 0 freezes Gain on the locked held-out ids. Later rounds call \texttt{evolve()} once; accept and rollback both score that same held-out. Held-out never enters \texttt{evolve()}. B0 is not ranked against B2.}
\label{fig:protocol}
\end{figure}

\texttt{no\_evolve} returns the same agent. Its four held-out passes are repeated measurements of a frozen policy~\citep{henderson2018matters,madaan2024variance}. The noise band is the max-minus-min of those four scores. Paired excess is a method's Gain minus the same model and seed's \texttt{no\_evolve} Gain. Excess inside the band is not reported as evolution. Runner eligibility (\texttt{main\_table\_eligible}) is a separate flag: a frozen \texttt{no\_evolve} cell can pass it while \texttt{agent\_version} stays $M_0$.

Paper-min, unless labeled otherwise: $R{=}3$; at least three seeds~\citep{bouthillier2021variance}; at least 40 evolution ids, 10 validation, 20 held-out; $M_0$ inside $[0.10,0.80]$ or the run is flagged; \texttt{human\_edits}${=}0$. The runner also flags a pack with no OOD slice. Coding, the primary scoreboard, is one family and has no OOD slice; that flag does not move the scoreboard. Each pack uses 80/20/40, so $n{=}40$ held-out (2.5 pp per task). Default mix \texttt{all}, schedule \texttt{accumulate\_chunks}, curriculum \texttt{none}~\citep{bengio2009curriculum}. Rank only inside $(\texttt{env},\texttt{budget\_class},\texttt{trace\_origin},\texttt{actor})$: B0 does not train weights and is not ranked with B2, a weight update. This paper reports B0 and B2 only; B1 is out of scope.

Table~\ref{tab:packs} is the pack contract.

\begin{table}[ht]
\caption{Three frozen packs. Each has 140 tasks, split 80/20/40, so held-out \(n{=}40\). The observation omits gold files, expected SQL, and unit-test oracles.}
\label{tab:packs}
\begin{center}
\scriptsize
\setlength{\tabcolsep}{3pt}
\begin{tabular}{@{}p{0.18\linewidth}p{0.26\linewidth}p{0.26\linewidth}p{0.24\linewidth}@{}}
\toprule
 & Office & Coding & Manufacturing \\
\midrule
Source & Product-usage workflows & 142 merged pull requests, rewritten as functions & 12 real tables; 31 real questions, 109 generated \\
Items & 10 families & One family, \texttt{fix\_bug} & 13 families \\
Outside evolution & 12/40: \texttt{access\_denied}, \texttt{do\_not\_publish} & None on held-out & 12/40: real questions on 4 unseen table families \\
Tools & \texttt{list\_files}, \texttt{read\_file}, \texttt{write\_file}, \texttt{publish\_artifact}, \texttt{grant\_access}, \texttt{submit} & \texttt{list\_dir}, \texttt{read\_file}, \texttt{write\_file}, \texttt{run\_tests}, \texttt{submit}; one file, \texttt{src/solution.py} & \texttt{list\_tables}, \texttt{describe\_table}, \texttt{execute\_sql}, \texttt{submit\_answer} \\
Success & Every terminal check passes & Hidden unit tests pass & Score $\ge 80$: SQL 40, result 40, answer 20 \\
Step cap & 20 & 12 & 12 \\
\bottomrule
\end{tabular}
\end{center}
\end{table}

\section{Coding measurement}
\label{sec:coding}

Coding is the primary scoreboard. It uses the same 80/20/40 split, $R{=}3$, and B0 Standard Tool Actor. The episode cap is 12 steps. Held-out is one family, \texttt{fix\_bug} ($n{=}40$). A task is one file, \texttt{src/solution.py}. The tools are \texttt{list\_dir}, \texttt{read\_file}, \texttt{write\_file}, \texttt{run\_tests}, and \texttt{submit}. Success is the hidden unit tests; the gold patch is not in the observation. Every miss is the 12-step cap, so a gain here does not open a second task type. The three methods, \texttt{gepa\_llm}, \texttt{ace\_llm}, and \texttt{expel\_llm}, were fixed on an office validation screen before held-out was read (Appendix~\ref{app:screen}). That screen is a separate smoke split: one round, six tasks, ten steps. Five methods passed the artifact check. The screen kept the three with the highest mean validation gain, then lower screen regression, then fewer selection tokens. Their mean screen gains were $0$, $-5.6$, and $-11.1$ pp, none positive (Table~\ref{tab:screenrank}). Those gains do not predict the held-out excess on this board. Each method reads that round's own traces and injects the text in Table~\ref{tab:slots} into the same frozen model.

\begin{table}[ht]
\caption{Screen ranking across three models at seed 42. Mean gain is in percentage points. Mean reg.\ is a rate. Valid is the artifact check.}
\label{tab:screenrank}
\begin{center}
\scriptsize
\setlength{\tabcolsep}{4pt}
\begin{tabular}{lrrrr}
\toprule
Method & Valid & Gain (pp) & Mean reg. & Tokens \\
\midrule
\texttt{gepa\_llm} & true & $0.0$ & 0.056 & 469231 \\
\texttt{ace\_llm} & true & $-5.6$ & 0.111 & 513033 \\
\texttt{expel\_llm} & true & $-11.1$ & 0.056 & 448950 \\
\texttt{dgm} & true & $-16.7$ & 0.111 & 396406 \\
\texttt{clin\_llm} & true & $-16.7$ & 0.222 & 411690 \\
\texttt{awm\_llm} & false & $0.0$ & 0.000 & 447296 \\
\texttt{reasoning\_bank} & false & $-11.1$ & 0.056 & 457843 \\
\bottomrule
\end{tabular}
\end{center}
\end{table}

\begin{table}[ht]
\caption{Text injected by the three scored methods. Each reads that round's own traces. The kept text is the TEB mechanism.}
\label{tab:slots}
\begin{center}
\scriptsize
\setlength{\tabcolsep}{4pt}
\begin{tabular}{lp{0.72\linewidth}}
\toprule
Method & Injected text \\
\midrule
\texttt{gepa\_llm} & One revised instruction. The validation gate accepts it or rolls it back. \\
\texttt{ace\_llm} & A grow-and-refine playbook from a generator, reflector, and curator. \\
\texttt{expel\_llm} & A deduplicated pool of insights from success and failure. \\
\bottomrule
\end{tabular}
\end{center}
\end{table}

DeepSeek \texttt{no\_evolve} keeps version $M_0$ with four-pass ranges $B=7.5$, $5.0$, and $12.5$ pp (Table~\ref{tab:coding}, Figure~\ref{fig:codingrounds}). Frozen Gain and band for all three models are Table~\ref{tab:codingfrozen}.

\begin{figure}[ht]
\centering
\includegraphics[width=\linewidth]{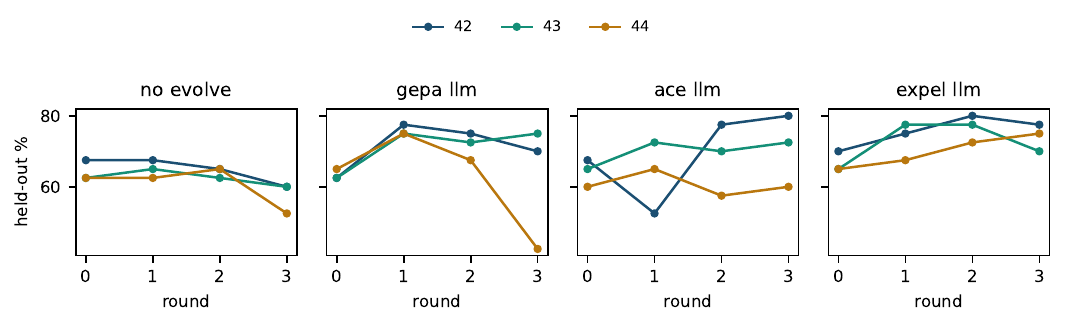}
\caption{DeepSeek coding held-out success by round. \texttt{no\_evolve} keeps version $M_0$ and the score still moves. Seeds are 42, 43, and 44.}
\label{fig:codingrounds}
\end{figure}

\begin{table}[ht]
\caption{Coding, DeepSeek v4 Pro, \texttt{no\_evolve} only. Success is held-out percent. Version is $M_0$ at every round. Range is that seed's band.}
\label{tab:coding}
\begin{center}
\scriptsize
\setlength{\tabcolsep}{4pt}
\begin{tabular}{rrrrrrr}
\toprule
Seed & r0 & r1 & r2 & r3 & Gain & Range \\
\midrule
42 & 67.5 & 67.5 & 65.0 & 60.0 & $-7.5$ & 7.5 \\
43 & 62.5 & 65.0 & 62.5 & 60.0 & $-2.5$ & 5.0 \\
44 & 62.5 & 62.5 & 65.0 & 52.5 & $-10.0$ & 12.5 \\
\bottomrule
\end{tabular}
\end{center}
\end{table}

Figure~\ref{fig:codingrounds} is why excess, not the raw curve, is the coding result. \texttt{no\_evolve} keeps version $M_0$ while each seed's curve still moves, and its Gain is negative on all three seeds. A method panel can rise while part of that rise is the frozen score falling.

\begin{table}[ht]
\caption{Coding \texttt{no\_evolve} by model. Frozen Gain is the four-pass Gain. Band is that seed's four-pass range. DeepSeek pass scores are Table~\ref{tab:coding}.}
\label{tab:codingfrozen}
\begin{center}
\scriptsize
\setlength{\tabcolsep}{4pt}
\begin{tabular}{llrr}
\toprule
Model & Seed & Frozen Gain & Band \\
\midrule
DeepSeek & 42 & $-7.5$ & 7.5 \\
DeepSeek & 43 & $-2.5$ & 5.0 \\
DeepSeek & 44 & $-10.0$ & 12.5 \\
GLM & 42 & $-2.5$ & 12.5 \\
GLM & 43 & $+7.5$ & 7.5 \\
GLM & 44 & $-7.5$ & 15.0 \\
MiniMax & 42 & $+5.0$ & 10.0 \\
MiniMax & 43 & $0.0$ & 2.5 \\
MiniMax & 44 & $0.0$ & 12.5 \\
\bottomrule
\end{tabular}
\end{center}
\end{table}

Table~\ref{tab:codingfrozen} is the cross-model control. DeepSeek is the only model whose three frozen Gains are all negative, so a rising method panel on that model includes the frozen score falling. GLM seeds 42 and 44 are also negative. GLM seed 44's band is $15.0$ pp, so an excess of $+15.0$ equals that band. The same $+15.0$ on DeepSeek seed 42 exceeds its $7.5$ pp band.

Table~\ref{tab:codingexcess} is DeepSeek paired excess. Seven of the nine cells are positive and larger than that seed's own $B$. All seven have final regression at or below $0.20$: \texttt{gepa\_llm} seeds 42 and 43 ($+15.0$ against bands $7.5$ and $5.0$), \texttt{ace\_llm} seeds 42 and 43 ($+20.0$ and $+10.0$), and \texttt{expel\_llm} seeds 42, 43, and 44 ($+15.0$, $+7.5$, $+20.0$). Round-0 success on these seven cells is within $5$ pp of the same-seed frozen round 0. \texttt{gepa\_llm} seed 44 moves the other way ($-12.5$, equal to its $12.5$ band) and its final regression is $0.48$. The cell stays in the excess comparison. One seed of the same method moves down by the width of its band. The seven cells above their bands leave version $M_0$. \texttt{ace\_llm} seed 44, the positive cell inside its band, keeps version $M_0$. Leaving version $M_0$ is not the excess: \texttt{gepa\_llm} seed 44 also finishes at version $M_2$, and its excess is $-12.5$.

\begin{table}[ht]
\caption{Coding-v1, DeepSeek v4 Pro. Excess is method Gain minus the same-seed \texttt{no\_evolve} Gain, in percentage points. Band is that seed's four-pass range. Reg.\ is the final regression rate. None of these nine cells is \texttt{over\_budget}.}
\label{tab:codingexcess}
\begin{center}
\scriptsize
\setlength{\tabcolsep}{3pt}
\begin{tabular}{llrrrrrl}
\toprule
Method & Seed & Gain & Frozen & Excess & Band & Reg. & Against that band \\
\midrule
gepa & 42 & $+7.5$ & $-7.5$ & $+15.0$ & 7.5 & 0.167 & outside \\
gepa & 43 & $+12.5$ & $-2.5$ & $+15.0$ & 5.0 & 0.069 & outside \\
gepa & 44 & $-22.5$ & $-10.0$ & $-12.5$ & 12.5 & 0.481 & equal, down \\
ace & 42 & $+12.5$ & $-7.5$ & $+20.0$ & 7.5 & 0.065 & outside \\
ace & 43 & $+7.5$ & $-2.5$ & $+10.0$ & 5.0 & 0.071 & outside \\
ace & 44 & $0.0$ & $-10.0$ & $+10.0$ & 12.5 & 0.130 & inside \\
expel & 42 & $+7.5$ & $-7.5$ & $+15.0$ & 7.5 & 0.062 & outside \\
expel & 43 & $+5.0$ & $-2.5$ & $+7.5$ & 5.0 & 0.097 & outside \\
expel & 44 & $+10.0$ & $-10.0$ & $+20.0$ & 12.5 & 0.103 & outside \\
\bottomrule
\end{tabular}
\end{center}
\end{table}

\begin{figure}[ht]
\centering
\includegraphics[width=\linewidth]{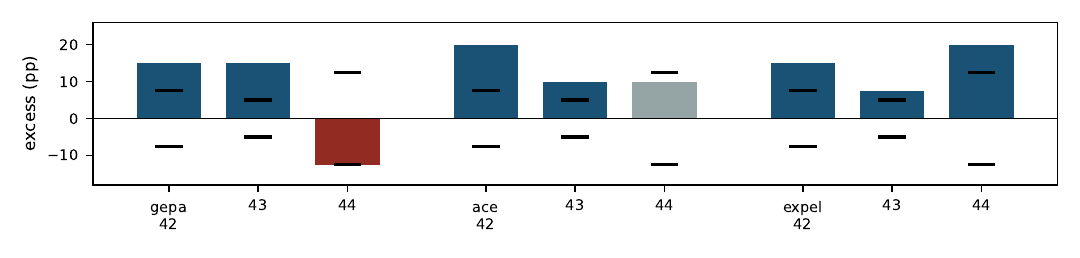}
\caption{DeepSeek coding paired excess. Black ticks are that seed's own band. Dark bars lie strictly above it. The light positive bar is \texttt{ace\_llm} seed 44, inside its $12.5$ pp band. \texttt{gepa\_llm} seed 44 equals its band on the way down.}
\label{fig:codingexcess}
\end{figure}

\ifdefined\tebarxiv\else
\begin{wrapfigure}{r}{2.55in}
\vspace{-0.8em}
\centering
\includegraphics[width=2.35in]{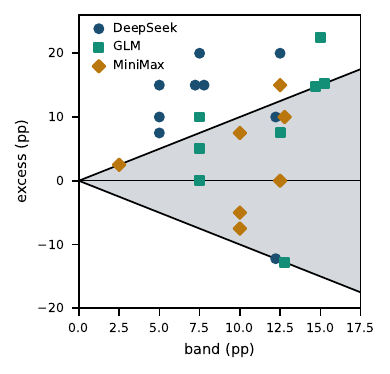}
\caption{Paired cells. The wedge is inside the band. A point above the upper line clears that seed's band. Unpaired cells are omitted.}
\label{fig:codingplane}
\vspace{-0.6em}
\end{wrapfigure}
\fi
GLM coding is the same comparison on a wider frozen range (Table~\ref{tab:codingstatus}). \texttt{gepa\_llm} excesses are $+7.5$, $0$, and $+15.0$ against bands $12.5$, $7.5$, and $15.0$: the first two sit inside, and seed 44 equals its band. \texttt{ace\_llm} seed 42 is $-12.5$ on a $12.5$ band and is \texttt{over\_budget}; seeds 43 and 44 are $+5.0$ (inside $7.5$) and $+15.0$ (equal to $15.0$), and seed 43 is also \texttt{over\_budget}. \texttt{expel\_llm} seeds 43 and 44 are $+10.0$ and $+22.5$ against bands $7.5$ and $15.0$, with final regression $0.03$ and $0.13$. Seed 44's round 0 is $60.0$ against the frozen round 0 of $67.5$, a $7.5$ pp gap, below the 10 pp cutoff, so that excess stays in the comparison. \texttt{expel\_llm} seed 42 is not in this pairing: its recorded $M_0$ is $5\%$, while \texttt{no\_evolve} seed 42 opens at $72.5\%$. The round-0 gap is $67.5$ pp. The cell is outside the paper-min $M_0$ window $[0.10,0.80]$.

MiniMax coding is finished for all three methods on seeds 42, 43, and 44. On seed 42 the excesses are $-7.5$, $-5.0$, and $+7.5$ for \texttt{gepa\_llm}, \texttt{ace\_llm}, and \texttt{expel\_llm}, all inside the $10.0$ band; \texttt{expel\_llm} has final regression $0.22$. Seed 43 \texttt{gepa\_llm} and \texttt{ace\_llm} open at $27.5\%$ against a frozen round 0 of $37.5\%$, so their excesses of $+45.0$ and $+17.5$ are not the same starting draw. \texttt{expel\_llm} seed 43 equals the $2.5$ band and has regression $0.36$. On seed 44, \texttt{gepa\_llm} Gain is $+10.0$ and its excess is $+10.0$, inside the $12.5$ band, with regression $0.14$; \texttt{ace\_llm} excess is $0$. \texttt{expel\_llm} excess is $+15.0$, above that $12.5$ band. Its round 0 is $37.5$ against the frozen round 0 of $40.0$, and its final regression is $0.30$.

\begin{table}[ht]
\caption{Coding status for GLM and MiniMax. Excess and band are defined as in Table~\ref{tab:codingexcess}. Unpaired means the round-0 gap is at least 10 pp. Reg.\ is the final regression rate. Cap marks \texttt{over\_budget}. A regression above $0.20$ is shown and does not remove the cell.}
\label{tab:codingstatus}
\begin{center}
\scriptsize
\setlength{\tabcolsep}{2.5pt}
\begin{tabular}{llrrrrrl}
\toprule
Model & Method & Seed & Excess & Band & Reg. & Cap & Against that band \\
\midrule
GLM & gepa & 42 & $+7.5$ & 12.5 & 0.207 & & inside \\
GLM & gepa & 43 & $0.0$ & 7.5 & 0.069 & & inside \\
GLM & gepa & 44 & $+15.0$ & 15.0 & 0.094 & & equal \\
GLM & ace & 42 & $-12.5$ & 12.5 & 0.130 & yes & equal, down \\
GLM & ace & 43 & $+5.0$ & 7.5 & 0.065 & yes & inside \\
GLM & ace & 44 & $+15.0$ & 15.0 & 0.091 & & equal \\
GLM & expel & 42 & $+70.0$ & 12.5 & 0.000 & & unpaired \\
GLM & expel & 43 & $+10.0$ & 7.5 & 0.031 & & outside \\
GLM & expel & 44 & $+22.5$ & 15.0 & 0.125 & & outside \\
MiniMax & gepa & 42 & $-7.5$ & 10.0 & 0.353 & & inside \\
MiniMax & gepa & 43 & $+45.0$ & 2.5 & 0.000 & & unpaired \\
MiniMax & gepa & 44 & $+10.0$ & 12.5 & 0.136 & & inside \\
MiniMax & ace & 42 & $-5.0$ & 10.0 & 0.214 & & inside \\
MiniMax & ace & 43 & $+17.5$ & 2.5 & 0.158 & & unpaired \\
MiniMax & ace & 44 & $0.0$ & 12.5 & 0.188 & & inside \\
MiniMax & expel & 42 & $+7.5$ & 10.0 & 0.222 & & inside \\
MiniMax & expel & 43 & $+2.5$ & 2.5 & 0.364 & & equal \\
MiniMax & expel & 44 & $+15.0$ & 12.5 & 0.304 & & outside \\
\bottomrule
\end{tabular}
\end{center}
\end{table}

\ifdefined\tebarxiv
\begin{figure}[ht]
\centering
\includegraphics[width=2.55in]{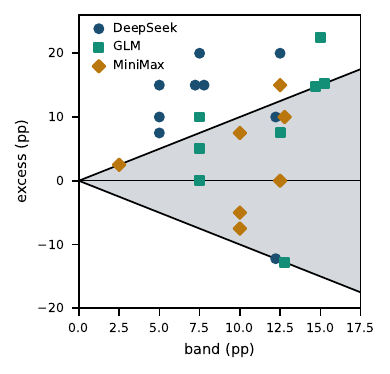}
\caption{Paired cells. The wedge is inside the band. A point above the upper line clears that seed's band. Unpaired cells are omitted.}
\label{fig:codingplane}
\end{figure}
\begin{figure}[ht]
\else
\begin{figure}[H]
\fi
\centering
\includegraphics[width=\linewidth]{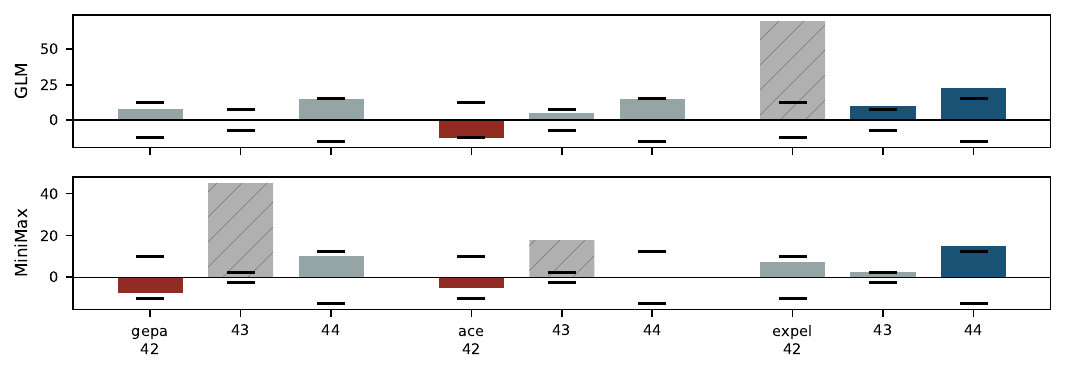}
\caption{GLM and MiniMax coding paired excess. Black ticks are that seed's own band. Dark bars lie strictly above it: GLM \texttt{expel\_llm} seeds 43 and 44, and MiniMax \texttt{expel\_llm} seed 44. Hatched bars are not the same starting draw: GLM \texttt{expel\_llm} seed 42, and MiniMax \texttt{gepa\_llm} and \texttt{ace\_llm} seed 43.}
\label{fig:codingexcessother}
\end{figure}

Figure~\ref{fig:codingexcessother} is why the seven-of-nine pattern stays a DeepSeek result. Three dark bars clear their own bands, not seven of nine. A hatched bar is a different starting draw and is not one of those seven. Figure~\ref{fig:codingplane} is the same paired cells. DeepSeek's seven cells lie above the upper boundary. GLM's $+15$ at seed 44 lies on the boundary of a $15$ pp band. DeepSeek's $+15$ at seed 42 lies above the boundary of a $7.5$ pp band.

\section{Manufacturing measurement}
\label{sec:mfgbody}

Manufacturing asks whether the coding seven-of-nine pattern appears for the same three methods. It does not. A pass lists tables, describes one, executes SQL, and submits an answer. Success is a deterministic score of at least 80: SQL 40, result 40, answer 20. The expected SQL is not in the observation. The pack uses the same 80/20/40 split and the 12-step cap. DeepSeek \texttt{no\_evolve} ranges are $7.5$, $5.0$, and $2.5$ pp. GLM ranges are $2.5$, $5.0$, and $2.5$. MiniMax ranges are $12.5$, $10.0$, and $12.5$. One task is still $2.5$ pp.

No finished DeepSeek manufacturing cell has a positive excess larger than its seed band (Table~\ref{tab:mfgexcess}). \texttt{ace\_llm} excess is $-2.5$ on every seed; Gain on those three cells is $0$. \texttt{expel\_llm} seed 44 is $+2.5$, equal to the $2.5$ band. \texttt{gepa\_llm} seeds 42 and 43 have excesses $-10.0$ and $-12.5$. Seed 42 \texttt{gepa\_llm} and seed 42 \texttt{expel\_llm} both open at $45.0$ against a frozen round 0 of $35.0$, so those two excesses are not the same starting draw. \texttt{gepa\_llm} seed 43 opens at $47.5$ against $40.0$, a $7.5$ pp gap, and stays in the comparison. \texttt{ace\_llm} seed 44 and \texttt{expel\_llm} seed 43 have final regression $0.21$, above $0.20$. GLM's largest positive excess is $+5.0$ (\texttt{ace\_llm} seed 43), equal to that seed's $5.0$ band. \texttt{expel\_llm} seed 42 is $-7.5$ against a $2.5$ band. The other GLM excesses are $0$ or $+2.5$ on bands of $2.5$ or $5.0$.

\begin{table}[ht]
\caption{Manufacturing-v1, DeepSeek v4 Pro. Excess and band are defined as in Table~\ref{tab:codingexcess}.}
\label{tab:mfgexcess}
\begin{center}
\scriptsize
\setlength{\tabcolsep}{4pt}
\begin{tabular}{llrrrrl}
\toprule
Method & Seed & Gain & Frozen & Excess & Band & Against that band \\
\midrule
gepa & 42 & $-7.5$ & $+2.5$ & $-10.0$ & 7.5 & outside, negative; r0 gap 10 pp \\
gepa & 43 & $-10.0$ & $+2.5$ & $-12.5$ & 5.0 & outside, negative; r0 gap 7.5 pp \\
gepa & 44 & $0.0$ & $+2.5$ & $-2.5$ & 2.5 & equal \\
ace & 42 & $0.0$ & $+2.5$ & $-2.5$ & 7.5 & inside \\
ace & 43 & $0.0$ & $+2.5$ & $-2.5$ & 5.0 & inside \\
ace & 44 & $0.0$ & $+2.5$ & $-2.5$ & 2.5 & equal; reg.\ 0.21 \\
expel & 42 & $-7.5$ & $+2.5$ & $-10.0$ & 7.5 & outside, negative; r0 gap 10 pp \\
expel & 43 & $+2.5$ & $+2.5$ & $0.0$ & 5.0 & inside; reg.\ 0.21 \\
expel & 44 & $+5.0$ & $+2.5$ & $+2.5$ & 2.5 & equal \\
\bottomrule
\end{tabular}
\end{center}
\end{table}

MiniMax manufacturing is finished for all three methods on seeds 42, 43, and 44. On seed 42, \texttt{ace\_llm} excess is $-15.0$ and \texttt{expel\_llm} excess is $-7.5$ against a $12.5$ band, so only \texttt{ace\_llm} falls outside, and it falls below. \texttt{gepa\_llm} opens at $27.5$ against a frozen round 0 of $17.5$, a $10$ pp gap, so its $-7.5$ excess is not the same starting draw. Seed 43 \texttt{gepa\_llm} excess is $+5.0$ and \texttt{expel\_llm} excess is $-5.0$, both inside the $10.0$ band. \texttt{ace\_llm} seed 43 excess is $+20.0$ against that band, but round 0 is $12.5$ against a frozen round 0 of $22.5$. On seed 44, \texttt{ace\_llm} excess is $+2.5$ and \texttt{expel\_llm} excess is $-7.5$, both inside the $12.5$ band; \texttt{expel\_llm} keeps version $M_0$. \texttt{gepa\_llm} seed 44 Gain is $+25.0$ and its excess is $+20.0$ against that band, but round 0 is $12.5$ against a frozen round 0 of $22.5$. Final regression on all nine MiniMax method cells is above $0.20$.

\section{Office measurement}
\label{sec:officectr}

Office-v1 uses the same 80/20/40 split and the B0 Standard Tool Actor. The episode cap is 20 steps. The tools are \texttt{list\_files}, \texttt{read\_file}, \texttt{write\_file}, \texttt{publish\_artifact}, \texttt{grant\_access}, and \texttt{submit}. Success is every terminal check passing. A failed \texttt{submit} returns the failed check names and leaves the episode open. Gold files are not in the observation. It is scored on its own band and is not averaged with coding. Actor, prompt, and held-out ids stay fixed, and the request temperature is 0. The hosted endpoint is still not deterministic, so repeated scoring of the same version-$M_0$ policy is a band~\citep{atil2024nondeterminism}. DeepSeek v4 Pro \texttt{no\_evolve} keeps version $M_0$. The three seed ranges are $17.5$, $7.5$, and $17.5$ pp, so the widest frozen swing is $17.5$ pp; one task is 2.5 points. GLM seed 42 spans $17.5$ pp. MiniMax seed 42 spans $5.0$ pp. GLM and MiniMax are one seed on this pack.

No DeepSeek office cell has a positive excess larger than its own seed band (Table~\ref{tab:v1excess}). Mean excess is $-2.5$ pp (\texttt{gepa\_llm}), $-3.3$ pp (\texttt{ace\_llm}), and $+4.2$ pp (\texttt{expel\_llm}), all inside the widest $17.5$ pp range. Seed 43's own band is $7.5$ pp: \texttt{gepa\_llm} is $-7.5$ and \texttt{ace\_llm} is $+7.5$, so both sit on that band rather than beyond it. \texttt{incident\_track}, \texttt{postmortem}, and \texttt{onboarding} stay at zero solved tasks on every DeepSeek pass. The solved-count changes that remain are in families the frozen policy already solves on some pass.

\begin{table}[ht]
\caption{Office-v1, DeepSeek v4 Pro. Excess and band are defined as in Table~\ref{tab:codingexcess}. Reg.\ is the final regression rate.}
\label{tab:v1excess}
\begin{center}
\scriptsize
\setlength{\tabcolsep}{3pt}
\begin{tabular}{llrrrrrl}
\toprule
Method & Seed & Gain & Frozen & Excess & Band & Reg. & Against that band \\
\midrule
gepa & 42 & $+7.5$ & $+10.0$ & $-2.5$ & 17.5 & 0.250 & inside \\
gepa & 43 & $0.0$ & $+7.5$ & $-7.5$ & 7.5 & 0.375 & equal \\
gepa & 44 & $+17.5$ & $+15.0$ & $+2.5$ & 17.5 & 0.143 & inside \\
ace & 42 & $+5.0$ & $+10.0$ & $-5.0$ & 17.5 & 0.267 & inside \\
ace & 43 & $+15.0$ & $+7.5$ & $+7.5$ & 7.5 & 0.077 & equal \\
ace & 44 & $+2.5$ & $+15.0$ & $-12.5$ & 17.5 & 0.385 & inside \\
expel & 42 & $+17.5$ & $+10.0$ & $+7.5$ & 17.5 & 0.154 & inside \\
expel & 43 & $+10.0$ & $+7.5$ & $+2.5$ & 7.5 & 0.000 & inside \\
expel & 44 & $+17.5$ & $+15.0$ & $+2.5$ & 17.5 & 0.091 & inside \\
\bottomrule
\end{tabular}
\end{center}
\end{table}

GLM excesses sit inside $17.5$ pp, and the three adapter cells are \texttt{over\_budget}. MiniMax \texttt{gepa\_llm} excess is $+2.5$ pp, inside the $5.0$ pp band. \texttt{ace\_llm} excess is $-7.5$ pp and stopped after round 2. \texttt{expel\_llm} excess is $+7.5$ pp, $2.5$ pp past that $5.0$ pp band, on one seed, at final version $M_1$. That is not a confirmed evolution result.

\section{Limitations}

Coding held-out is one family, \texttt{fix\_bug} ($n{=}40$, 2.5 pp per task). The seven cells above their own bands do not include \texttt{gepa\_llm} seed 44, which moves down by $12.5$ pp with final regression $0.48$. GLM and MiniMax do not reproduce that seven-of-nine pattern. GLM \texttt{expel\_llm} seeds 43 and 44 still lie above their own bands; seed 42 is unpaired ($M_0$ $5\%$ against a frozen round 0 of $72.5\%$). MiniMax \texttt{expel\_llm} seed 44 also lies above its $12.5$ band, with final regression $0.30$. Frozen coding seeds 42 and 44 have regression $0.231$ and $0.308$ with no policy change.

Office-v1 held-out has $n{=}40$. Effects under the 17.5 pp DeepSeek band are unresolved by this design~\citep{dror2018significance,card2020power}, and the $0.20$ regression cutoff already raises the runner's flag on frozen seed 43 ($0.231$, $M_0$). Validation has $n{=}20$ (5 pp per task). On office-v1, MiniMax and GLM are one seed. GLM adapters and MiniMax \texttt{ace\_llm} crossed the evolution-token cap. \texttt{awm\_llm}, \texttt{reasoning\_bank}, \texttt{clin\_llm}, and \texttt{dgm} were screened and are not on the office scoreboard. Closed hosted models were not weight-updated; TRACE and a B2 open-weight card~\citep{hu2022lora} are a different budget class.

Archived ticket-gym, inbox, and NL2SQL cards use another tool set and $n{=}20$ held-out and are not office results (Appendix~\ref{sec:office}).

Manufacturing is not a coding ranking (Section~\ref{sec:mfgbody}). Office uses 20 steps per task; coding and manufacturing use 12. B0 token costs outside the evolution-token cap are estimates. MiniMax manufacturing is complete on three seeds. The two MiniMax excesses outside their seed bands (\texttt{ace\_llm} seed 43 and \texttt{gepa\_llm} seed 44) are not the same starting draw, and all nine MiniMax method cells have final regression above $0.20$.

\section{Conclusion}

TEB's main result is attribution, not a single global winner. On coding, endpoint Gain includes frozen rescoring variance: DeepSeek \texttt{no\_evolve} keeps version $M_0$ while the three seeds span $7.5$, $5.0$, and $12.5$ pp (Section~\ref{sec:coding}). After seed-matched paired excess, seven of nine DeepSeek text-update cells lie above their own bands, but this pattern does not reproduce on GLM and MiniMax. Across all three models, no method stays above its own band on every seed (Table~\ref{tab:codingstatus}).

This separation between observed Gain and attributable excess is the core contribution: TEB turns self-evolution claims into a falsifiable protocol with explicit controls, rather than a single endpoint delta. The same contract keeps office and manufacturing as independent control packs on their own bands. Office DeepSeek does not exceed its own bands (widest $17.5$ pp; Section~\ref{sec:officectr}), and manufacturing has no paired positive excess beyond its seed band.

\section*{AI use statement}

Generative AI coding assistants participated in the early execution of the implementation, namely the TEB runner and the method adapters. They also participated in polishing some prose, in retrieval, and in drafting some sections. The experimental design was set by the authors. We have not used generative AI to generate synthetic evaluation packs, formulate mathematical claims, write proofs, or invent held-out scores. All Success~\%, Gain, family counts, and gate decisions are read from stored run artifacts. The authors reviewed that implementation and take responsibility for the final content of this work.

\section*{Ethics statement}

Manufacturing-v1 derives from internal business tables provided by a manufacturing partner. The benchmark replaces counterparty, company, department, cost-center, and document identifiers with pseudonyms and withholds the private mapping. The tables contain no personal names, phone numbers, e-mail addresses, or identity numbers. Office and coding tasks are rewritten from product workflows and pull requests; contact details in office files use reserved example domains. No human subjects were involved.

\section*{Reproducibility statement}

Section~\ref{sec:protocol} defines every scored quantity. Appendix~\ref{app:adapters} lists what each adapter keeps, Appendix~\ref{app:screen} the development screen, Appendix~\ref{app:flags} the eligibility flags, and Appendix~\ref{app:examples} one task from each suite. The release contains the runner and table contract (\texttt{teb/}, \path{teb/table_contract.py}), the three packs with source metadata removed (\texttt{tasks/office}, \texttt{tasks/coding}, \texttt{tasks/manufacturing}), the configs of the reported runs, and the sanitized per-run summaries used by \path{scripts/plot_office_v1.py}. Appendix~\ref{app:data} provides the dataset card and licenses (data CC BY 4.0, code MIT). Hosted endpoints require the reader's own API key.

\bibliography{refs,refs_added}
\bibliographystyle{plainnat}

\appendix
\renewcommand{\thesection}{A.\arabic{section}}
\startcontents[app]
\section*{Appendix contents}
\begingroup
\makeatletter
\renewcommand*\l@section{\@dottedtocline{1}{0pt}{3.6em}}
\renewcommand*\l@subsection{\@dottedtocline{2}{3.6em}{5.4em}}
\makeatother
\printcontents[app]{}{1}{\setcounter{tocdepth}{2}}
\endgroup

\section{Dataset card}
\label{app:data}

Following datasheet practice~\citep{gebru2021datasheets}, TEB ships three frozen packs with the same 80/20/40 split and fixed held-out membership (Table~\ref{tab:datacard}). Each pack is adapted from a real source: product-usage workflows for office, merged pull requests of one production repository for coding, and one manufacturer's internal business tables for manufacturing. Manufacturing contains 31 real questions; the other 109 are generated over the same real tables with recomputed gold, so ``real-derived'' refers to the tables and those 31 questions, not every item. The generated questions come from deterministic templates whose gold is computed by executing SQL on the released tables; a candidate is rejected if it repeats an original question, gold SQL, or answer value, names a real entity, or touches a held-out-only table. Coding tasks restate each pull request's behavior as a rewritten Python function; no repository code is copied. Two held-out sets contain items outside the evolution distribution: office adds two held-out-only families, and manufacturing places its 12 held-out real questions on four table families that never appear in evolution or validation. A frozen DeepSeek v4 Pro policy leaves headroom on every pack, and five of the 13 manufacturing families are unsolved on every pass (Appendix~\ref{app:mfgfam}). Gold files, expected SQL, and unit-test oracles never enter the observation. The packs are not claimed to measure production performance or open-ended NL2SQL generalization~\citep{yu2018spider,li2023bird}; manufacturing scoring checks execution rather than string match~\citep{zhong2020sqlsuite}. Verifiers are deterministic. The reference solution of every one of the 420 items (gold files, gold patch, or gold SQL) passes its verifier, and the released tests check this. We did not conduct a separate annotator audit. Benchmark data are licensed under CC BY 4.0 and code under MIT.

\begin{table}[ht]
\caption{TEB pack card. Frozen range is held-out Success over the four passes of DeepSeek v4 Pro \texttt{no\_evolve}, seeds 42--44. Language counts instructions containing Chinese.}
\label{tab:datacard}
\begin{center}
\scriptsize
\begin{tabularx}{\linewidth}{@{}>{\raggedright\arraybackslash}p{0.15\linewidth}>{\raggedright\arraybackslash}X>{\raggedright\arraybackslash}X>{\raggedright\arraybackslash}X@{}}
\toprule
 & office-v1 & coding-v1 & manufacturing-v1 \\
\midrule
Source & Product-usage workflows in a closed catalog workspace & 142 merged pull requests of one production repository & 12 pseudonymized real tables (32,290 rows) and 31 real questions \\
Items & 140; 10 families & 140; \texttt{fix\_bug} & 140; 13 families; 31 real, 109 generated \\
Held-out OOD & 12/40: \texttt{access\_denied}, \texttt{do\_not\_publish} & none & 12/40: real questions on 4 unseen table families \\
Language & 33/140 Chinese, rest English & Chinese & Chinese \\
Observation & Catalog files and six tools & One file, \texttt{src/solution.py} & Four tools: list tables, describe a table, execute SQL, submit an answer \\
Success & Every terminal check passes & Closed-workspace unit tests pass & Score $\ge 80$ of 100: SQL 40, result 40, answer 20; a partial SQL or result check caps the score below 80 \\
Step cap & 20 & 12 & 12 \\
Frozen range & 22.5--45.0\% & 52.5--67.5\% & 35.0--45.0\% \\
\bottomrule
\end{tabularx}
\end{center}
\end{table}
\section{What each office adapter keeps}
\label{app:adapters}

All seven adapters read the current round's own traces, update once before the validation gate, and inject bounded text into the same frozen model. None updates weights. \texttt{no\_evolve} returns the same agent and still runs the round schedule, so the baseline is round-matched. The slots are method-inspired by ExpeL~\citep{zhao2024expel}, workflow memory~\citep{wang2024awm}, ACE~\citep{zhang2026ace}, GEPA~\citep{agrawal2025gepa}, ReasoningBank~\citep{ouyang2026reasoningbank}, CLIN~\citep{majumder2024clin}, and DGM~\citep{zhang2025dgm}. Upstream benchmark numbers are not reused.

\begin{table}[ht]
\caption{Method-inspired slots on the Standard Tool Actor. The kept column is the TEB mechanism. The omitted column is the upstream system this paper does not claim to reproduce.}
\label{tab:adapters}
\begin{center}
\scriptsize
\begin{tabular}{p{0.18\linewidth}p{0.38\linewidth}p{0.34\linewidth}}
\toprule
Slot & Kept & Omitted \\
\midrule
\texttt{expel\_llm} & Insights from success and failure; deduplicated pool & Upstream harness and retrieval \\
\texttt{awm\_llm} & Workflows from successful traces only & Upstream induction stack \\
\texttt{ace\_llm} & Generator, reflector, curator; grow-and-refine playbook & Upstream application runtime \\
\texttt{gepa\_llm} & One revised instruction; the TEB gate accepts or rolls it back & Population search and Pareto compile \\
\texttt{reasoning\_bank} & Title, when, and content memories; lexical top-$k$ & Upstream agent and test-time search \\
\texttt{clin\_llm} & Condition, action, and effect rules & Domain gold memory \\
\texttt{dgm} & Archive, parent score, text mutation & Repository self-modification \\
\bottomrule
\end{tabular}
\end{center}
\end{table}

\texttt{dgm\_native\_code} is a separate board (Appendix~\ref{app:native}). AgentEvolver~\citep{zhai2025agentevolver}, the SEED context probe, and TRACE~\citep{kang2026trace} are not part of Table~\ref{tab:v1grid}. TRACE remains a B2 card: contrast traces, synthesize environments, train LoRA experts~\citep{hu2022lora}. A closed hosted model cannot take that update.
\section{Development screen}
\label{app:screen}
The screen file records \texttt{held\_out\_read=false}. Gain in this section is validation success after one update minus the model-matched \texttt{no\_evolve} validation success on the same screen pass. Screen regression is $\max(0,\mathrm{before}-\mathrm{after})$ on that validation pass. It is not the held-out regression rate in Table~\ref{tab:v1excess}. Invalid rows failed the artifact check: the evolve artifact had to carry a hash, a non-empty context state, and \texttt{implementation\_fidelity=method-inspired}. The three valid methods with the highest mean gain, then lower screen regression, then fewer selection tokens, were extended to seeds 43 and 44. Those means are $0$, $-0.056$, and $-0.111$ for \texttt{gepa\_llm}, \texttt{ace\_llm}, and \texttt{expel\_llm}: $0$, $-5.6$, and $-11.1$ pp. None is a positive validation gain. The screen budget is one round, 10 steps per task, and 6 tasks per round, with \texttt{pack\_role=smoke} on a separate split. Formal office-v1 is $R{=}3$, 20 steps, and 80/20/40. Selection used the shorter episode. This screen is an office validation screen. The three methods it fixed are the ones scored on the coding primary scoreboard; that board was not a separate coding pre-registration (Section~\ref{sec:coding}). The ranking is Table~\ref{tab:screenrank}.
\begin{longtable}{llrrrrl}
\caption{Screen cells. Validation columns are success rates. Gain is after minus the model-matched \texttt{no\_evolve} rate.}\label{tab:screencells}\\
\toprule Model & Method & Before & After & Gain & Reg. & Artifact \\
\midrule\endfirsthead
\toprule Model & Method & Before & After & Gain & Reg. & Artifact \\
\midrule\endhead
\bottomrule\endfoot
deepseek-v4-pro & expel\_llm & 0.167 & 0.167 & -0.333 & 0.000 & true \\
deepseek-v4-pro & awm\_llm & 0.333 & 0.333 & -0.167 & 0.000 & true \\
deepseek-v4-pro & ace\_llm & 0.167 & 0.333 & -0.167 & 0.000 & true \\
deepseek-v4-pro & gepa\_llm & 0.500 & 0.333 & -0.167 & 0.167 & true \\
deepseek-v4-pro & reasoning\_bank & 0.333 & 0.167 & -0.333 & 0.167 & false \\
deepseek-v4-pro & clin\_llm & 0.500 & 0.333 & -0.167 & 0.167 & true \\
deepseek-v4-pro & dgm & 0.167 & 0.167 & -0.333 & 0.000 & true \\
minimax-m2-5 & expel\_llm & 0.167 & 0.167 & +0.000 & 0.000 & true \\
minimax-m2-5 & awm\_llm & 0.000 & 0.167 & +0.000 & 0.000 & false \\
minimax-m2-5 & ace\_llm & 0.333 & 0.167 & +0.000 & 0.167 & true \\
minimax-m2-5 & gepa\_llm & 0.000 & 0.167 & +0.000 & 0.000 & true \\
minimax-m2-5 & reasoning\_bank & 0.000 & 0.000 & -0.167 & 0.000 & true \\
minimax-m2-5 & clin\_llm & 0.000 & 0.167 & +0.000 & 0.000 & true \\
minimax-m2-5 & dgm & 0.167 & 0.000 & -0.167 & 0.167 & true \\
glm-5-3 & expel\_llm & 0.500 & 0.333 & +0.000 & 0.167 & true \\
glm-5-3 & awm\_llm & 0.500 & 0.500 & +0.167 & 0.000 & false \\
glm-5-3 & ace\_llm & 0.500 & 0.333 & +0.000 & 0.167 & true \\
glm-5-3 & gepa\_llm & 0.500 & 0.500 & +0.167 & 0.000 & true \\
glm-5-3 & reasoning\_bank & 0.333 & 0.500 & +0.167 & 0.000 & true \\
glm-5-3 & clin\_llm & 0.500 & 0.000 & -0.333 & 0.500 & true \\
glm-5-3 & dgm & 0.500 & 0.333 & +0.000 & 0.167 & true \\
\end{longtable}
\section{Office validation gates}
\label{app:gatesv1}
The gate compares validation success of the current agent with the proposal. Both numbers are one pass on $n{=}20$ validation tasks. A drop rolls the proposal back. Table~\ref{tab:v1gate} is every non-zero round in the 20-cell matrix. Round 0 has no proposal.
\begin{table}[ht]\caption{Validation gate on office-v1. Columns are validation percent before and after the proposal. $n{=}20$, so one task is 5 pp.}\label{tab:v1gate}\begin{center}\scriptsize\setlength{\tabcolsep}{3pt}
\begin{tabular}{llrrrrrr}\toprule Model & Method & Seed & Round & Version & Before & After & Decision \\\midrule
DS & ace & 42 & 1 & $M1$ & 20.0 & 40.0 & accept \\
DS & ace & 42 & 2 & $M2$ & 35.0 & 50.0 & accept \\
DS & ace & 42 & 3 & $M3$ & 35.0 & 35.0 & accept \\
DS & ace & 43 & 1 & $M0$ & 30.0 & 20.0 & rollback \\
DS & ace & 43 & 2 & $M0$ & 25.0 & 5.0 & rollback \\
DS & ace & 43 & 3 & $M1$ & 15.0 & 35.0 & accept \\
DS & ace & 44 & 1 & $M1$ & 25.0 & 30.0 & accept \\
DS & ace & 44 & 2 & $M2$ & 25.0 & 35.0 & accept \\
DS & ace & 44 & 3 & $M2$ & 35.0 & 25.0 & rollback \\
DS & expel & 42 & 1 & $M1$ & 15.0 & 20.0 & accept \\
DS & expel & 42 & 2 & $M2$ & 20.0 & 35.0 & accept \\
DS & expel & 42 & 3 & $M3$ & 25.0 & 35.0 & accept \\
DS & expel & 43 & 1 & $M1$ & 25.0 & 35.0 & accept \\
DS & expel & 43 & 2 & $M2$ & 20.0 & 20.0 & accept \\
DS & expel & 43 & 3 & $M2$ & 25.0 & 20.0 & rollback \\
DS & expel & 44 & 1 & $M1$ & 15.0 & 25.0 & accept \\
DS & expel & 44 & 2 & $M2$ & 15.0 & 15.0 & accept \\
DS & expel & 44 & 3 & $M3$ & 30.0 & 30.0 & accept \\
DS & gepa & 42 & 1 & $M0$ & 25.0 & 15.0 & rollback \\
DS & gepa & 42 & 2 & $M1$ & 20.0 & 30.0 & accept \\
DS & gepa & 42 & 3 & $M1$ & 35.0 & 25.0 & rollback \\
DS & gepa & 43 & 1 & $M0$ & 30.0 & 25.0 & rollback \\
DS & gepa & 43 & 2 & $M1$ & 20.0 & 35.0 & accept \\
DS & gepa & 43 & 3 & $M1$ & 20.0 & 5.0 & rollback \\
DS & gepa & 44 & 1 & $M0$ & 30.0 & 25.0 & rollback \\
DS & gepa & 44 & 2 & $M0$ & 20.0 & 10.0 & rollback \\
DS & gepa & 44 & 3 & $M1$ & 20.0 & 25.0 & accept \\
DS & no\_evolve & 42 & 1 & $M0$ & 20.0 & 15.0 & rollback \\
DS & no\_evolve & 42 & 2 & $M0$ & 20.0 & 20.0 & accept \\
DS & no\_evolve & 42 & 3 & $M0$ & 25.0 & 25.0 & accept \\
DS & no\_evolve & 43 & 1 & $M0$ & 30.0 & 15.0 & rollback \\
DS & no\_evolve & 43 & 2 & $M0$ & 30.0 & 15.0 & rollback \\
DS & no\_evolve & 43 & 3 & $M0$ & 20.0 & 20.0 & accept \\
DS & no\_evolve & 44 & 1 & $M0$ & 30.0 & 15.0 & rollback \\
DS & no\_evolve & 44 & 2 & $M0$ & 20.0 & 25.0 & accept \\
DS & no\_evolve & 44 & 3 & $M0$ & 30.0 & 25.0 & rollback \\
GLM & ace & 42 & 1 & $M0$ & 50.0 & 40.0 & rollback \\
GLM & ace & 42 & 2 & $M0$ & 45.0 & 35.0 & rollback \\
GLM & ace & 42 & 3 & $M0$ & --- & --- & over budget \\
GLM & expel & 42 & 1 & $M0$ & 50.0 & 35.0 & rollback \\
GLM & expel & 42 & 2 & $M0$ & 60.0 & 30.0 & rollback \\
GLM & expel & 42 & 3 & $M0$ & --- & --- & over budget \\
GLM & gepa & 42 & 1 & $M1$ & 50.0 & 50.0 & accept \\
GLM & gepa & 42 & 2 & $M1$ & 50.0 & 45.0 & rollback \\
GLM & gepa & 42 & 3 & $M1$ & --- & --- & over budget \\
GLM & no\_evolve & 42 & 1 & $M0$ & 45.0 & 40.0 & rollback \\
GLM & no\_evolve & 42 & 2 & $M0$ & 60.0 & 45.0 & rollback \\
GLM & no\_evolve & 42 & 3 & $M0$ & 55.0 & 30.0 & rollback \\
MM & ace & 42 & 1 & $M1$ & 10.0 & 15.0 & accept \\
MM & ace & 42 & 2 & $M1$ & --- & --- & over budget \\
MM & expel & 42 & 1 & $M0$ & 10.0 & 5.0 & rollback \\
MM & expel & 42 & 2 & $M0$ & 15.0 & 10.0 & rollback \\
MM & expel & 42 & 3 & $M1$ & 0.0 & 5.0 & accept \\
MM & gepa & 42 & 1 & $M1$ & 5.0 & 5.0 & accept \\
MM & gepa & 42 & 2 & $M2$ & 5.0 & 5.0 & accept \\
MM & gepa & 42 & 3 & $M3$ & 10.0 & 10.0 & accept \\
MM & no\_evolve & 42 & 1 & $M0$ & 10.0 & 5.0 & rollback \\
MM & no\_evolve & 42 & 2 & $M0$ & 10.0 & 10.0 & accept \\
MM & no\_evolve & 42 & 3 & $M0$ & 5.0 & 0.0 & rollback \\
\bottomrule\end{tabular}\end{center}\end{table}
One validation task is 5 pp. Several accepts in Table~\ref{tab:v1gate} are $0$ pp changes, which \texttt{min\_delta}${=}0$ keeps. On \texttt{no\_evolve} the artifact stays $M_0$ whether the row says accept or rollback: the two validation numbers are two passes of one frozen policy. GLM \texttt{no\_evolve} validation moves $45{\to}40$, $60{\to}45$, and $55{\to}30$. A later held-out move on that same version is a rescore, as in Table~\ref{tab:v1jump}.
\section{Per-round held-out record}
\label{app:rounds}
Table~\ref{tab:rounds} is every held-out pass in the 20-cell matrix. Regression is against the previous pass; round 0 is 0. Tokens are \texttt{total\_tokens} summed on that pass's held-out traces, not the 500{,}000 evolution-token cap. Validation is the stored \texttt{val\_success}. Main-text regression rates are these values rounded to two decimals.
\begin{longtable}{llrrrrrr}
\caption{Office-v1 held-out record. Success and validation are percent. Regression is a rate.}\label{tab:rounds}\\
\toprule Model & Method & Seed & Round & Succ. & Val. & Reg. & Tokens \\
\midrule\endfirsthead
\toprule Model & Method & Seed & Round & Succ. & Val. & Reg. & Tokens \\
\midrule\endhead
\bottomrule\endfoot
DS & ace & 42 & 0 & 32.5 & --- & 0.000 & 592020 \\
DS & ace & 42 & 1 & 30.0 & 40.0 & 0.231 & 690436 \\
DS & ace & 42 & 2 & 37.5 & 50.0 & 0.083 & 667912 \\
DS & ace & 42 & 3 & 37.5 & 35.0 & 0.267 & 808663 \\
DS & ace & 43 & 0 & 32.5 & --- & 0.000 & 576054 \\
DS & ace & 43 & 1 & 25.0 & 30.0 & 0.308 & 634816 \\
DS & ace & 43 & 2 & 32.5 & 25.0 & 0.100 & 585464 \\
DS & ace & 43 & 3 & 47.5 & 35.0 & 0.077 & 958826 \\
DS & ace & 44 & 0 & 30.0 & --- & 0.000 & 579534 \\
DS & ace & 44 & 1 & 27.5 & 30.0 & 0.167 & 594075 \\
DS & ace & 44 & 2 & 32.5 & 35.0 & 0.364 & 679860 \\
DS & ace & 44 & 3 & 32.5 & 35.0 & 0.385 & 967211 \\
DS & expel & 42 & 0 & 25.0 & --- & 0.000 & 576918 \\
DS & expel & 42 & 1 & 32.5 & 20.0 & 0.200 & 704342 \\
DS & expel & 42 & 2 & 32.5 & 35.0 & 0.308 & 786634 \\
DS & expel & 42 & 3 & 42.5 & 35.0 & 0.154 & 776706 \\
DS & expel & 43 & 0 & 32.5 & --- & 0.000 & 562761 \\
DS & expel & 43 & 1 & 30.0 & 35.0 & 0.077 & 671670 \\
DS & expel & 43 & 2 & 20.0 & 20.0 & 0.417 & 773513 \\
DS & expel & 43 & 3 & 42.5 & 25.0 & 0.000 & 673400 \\
DS & expel & 44 & 0 & 32.5 & --- & 0.000 & 568396 \\
DS & expel & 44 & 1 & 27.5 & 25.0 & 0.231 & 680503 \\
DS & expel & 44 & 2 & 27.5 & 15.0 & 0.364 & 718343 \\
DS & expel & 44 & 3 & 50.0 & 30.0 & 0.091 & 694999 \\
DS & gepa & 42 & 0 & 35.0 & --- & 0.000 & 580628 \\
DS & gepa & 42 & 1 & 25.0 & 25.0 & 0.357 & 590063 \\
DS & gepa & 42 & 2 & 40.0 & 30.0 & 0.200 & 685235 \\
DS & gepa & 42 & 3 & 42.5 & 35.0 & 0.250 & 635933 \\
DS & gepa & 43 & 0 & 30.0 & --- & 0.000 & 571489 \\
DS & gepa & 43 & 1 & 30.0 & 30.0 & 0.167 & 555353 \\
DS & gepa & 43 & 2 & 40.0 & 35.0 & 0.167 & 692592 \\
DS & gepa & 43 & 3 & 30.0 & 20.0 & 0.375 & 736969 \\
DS & gepa & 44 & 0 & 25.0 & --- & 0.000 & 608887 \\
DS & gepa & 44 & 1 & 35.0 & 30.0 & 0.100 & 556708 \\
DS & gepa & 44 & 2 & 35.0 & 20.0 & 0.071 & 578150 \\
DS & gepa & 44 & 3 & 42.5 & 25.0 & 0.143 & 609770 \\
DS & no\_evolve & 42 & 0 & 35.0 & --- & 0.000 & 563887 \\
DS & no\_evolve & 42 & 1 & 32.5 & 20.0 & 0.286 & 577280 \\
DS & no\_evolve & 42 & 2 & 27.5 & 20.0 & 0.308 & 619302 \\
DS & no\_evolve & 42 & 3 & 45.0 & 25.0 & 0.091 & 511676 \\
DS & no\_evolve & 43 & 0 & 27.5 & --- & 0.000 & 611006 \\
DS & no\_evolve & 43 & 1 & 35.0 & 30.0 & 0.182 & 575530 \\
DS & no\_evolve & 43 & 2 & 32.5 & 30.0 & 0.143 & 598822 \\
DS & no\_evolve & 43 & 3 & 35.0 & 20.0 & 0.231 & 564660 \\
DS & no\_evolve & 44 & 0 & 25.0 & --- & 0.000 & 612008 \\
DS & no\_evolve & 44 & 1 & 30.0 & 30.0 & 0.200 & 597854 \\
DS & no\_evolve & 44 & 2 & 22.5 & 25.0 & 0.333 & 616784 \\
DS & no\_evolve & 44 & 3 & 40.0 & 30.0 & 0.000 & 534010 \\
GLM & ace & 42 & 0 & 47.5 & --- & 0.000 & 457328 \\
GLM & ace & 42 & 1 & 47.5 & 50.0 & 0.105 & 431379 \\
GLM & ace & 42 & 2 & 35.0 & 45.0 & 0.316 & 695475 \\
GLM & ace & 42 & 3 & 37.5 & --- & 0.071 & 689266 \\
GLM & expel & 42 & 0 & 37.5 & --- & 0.000 & 466603 \\
GLM & expel & 42 & 1 & 50.0 & 50.0 & 0.133 & 415846 \\
GLM & expel & 42 & 2 & 35.0 & 60.0 & 0.400 & 677535 \\
GLM & expel & 42 & 3 & 37.5 & --- & 0.071 & 698951 \\
GLM & gepa & 42 & 0 & 47.5 & --- & 0.000 & 433726 \\
GLM & gepa & 42 & 1 & 45.0 & 50.0 & 0.211 & 582084 \\
GLM & gepa & 42 & 2 & 37.5 & 50.0 & 0.444 & 1142477 \\
GLM & gepa & 42 & 3 & 32.5 & --- & 0.200 & 1229120 \\
GLM & no\_evolve & 42 & 0 & 42.5 & --- & 0.000 & 443087 \\
GLM & no\_evolve & 42 & 1 & 35.0 & 45.0 & 0.235 & 464064 \\
GLM & no\_evolve & 42 & 2 & 52.5 & 60.0 & 0.071 & 441988 \\
GLM & no\_evolve & 42 & 3 & 35.0 & 55.0 & 0.381 & 659535 \\
MM & ace & 42 & 0 & 32.5 & --- & 0.000 & 510135 \\
MM & ace & 42 & 1 & 20.0 & 15.0 & 0.538 & 775021 \\
MM & ace & 42 & 2 & 27.5 & --- & 0.000 & 765450 \\
MM & expel & 42 & 0 & 22.5 & --- & 0.000 & 523694 \\
MM & expel & 42 & 1 & 22.5 & 10.0 & 0.333 & 540754 \\
MM & expel & 42 & 2 & 20.0 & 15.0 & 0.222 & 533679 \\
MM & expel & 42 & 3 & 32.5 & 5.0 & 0.125 & 770089 \\
MM & gepa & 42 & 0 & 17.5 & --- & 0.000 & 524370 \\
MM & gepa & 42 & 1 & 20.0 & 5.0 & 0.286 & 565039 \\
MM & gepa & 42 & 2 & 27.5 & 5.0 & 0.375 & 537457 \\
MM & gepa & 42 & 3 & 22.5 & 10.0 & 0.273 & 533844 \\
MM & no\_evolve & 42 & 0 & 20.0 & --- & 0.000 & 527353 \\
MM & no\_evolve & 42 & 1 & 25.0 & 10.0 & 0.250 & 523358 \\
MM & no\_evolve & 42 & 2 & 20.0 & 10.0 & 0.300 & 525339 \\
MM & no\_evolve & 42 & 3 & 22.5 & 5.0 & 0.250 & 533712 \\
\end{longtable}
\section{DeepSeek family counts on every pass}
\label{app:famcount}
Counts are solved held-out tasks. Column order is \texttt{catalog\_discover} ($n{=}8$), \texttt{brief\_report} ($5$), \texttt{checklist} ($5$), \texttt{csv\_audit} ($4$), \texttt{access\_denied} ($6$), \texttt{do\_not\_publish} ($6$). \texttt{incident\_track}, \texttt{postmortem}, and \texttt{onboarding} are omitted: they are $0$ on every DeepSeek pass.
\begin{table}[ht]\caption{DeepSeek \texttt{no\_evolve} solved counts by family and round. \texttt{incident\_track}, \texttt{postmortem}, and \texttt{onboarding} are omitted (0 on every pass).}\label{tab:famno_evolve}\begin{center}\scriptsize\setlength{\tabcolsep}{3pt}
\begin{tabular}{rrrrrrrr}\toprule Seed & Round & \rotatebox{65}{\texttt{catalog\_discover}} & \rotatebox{65}{\texttt{brief\_report}} & \rotatebox{65}{\texttt{checklist}} & \rotatebox{65}{\texttt{csv\_audit}} & \rotatebox{65}{\texttt{access\_denied}} & \rotatebox{65}{\texttt{do\_not\_publish}} \\\midrule
42 & 0 & 6 & 0 & 0 & 1 & 2 & 5 \\
42 & 1 & 4 & 1 & 1 & 1 & 2 & 4 \\
42 & 2 & 4 & 0 & 0 & 1 & 1 & 5 \\
42 & 3 & 7 & 2 & 2 & 2 & 1 & 4 \\
43 & 0 & 5 & 1 & 0 & 1 & 1 & 3 \\
43 & 1 & 6 & 0 & 1 & 1 & 2 & 4 \\
43 & 2 & 5 & 0 & 1 & 1 & 2 & 4 \\
43 & 3 & 5 & 2 & 1 & 1 & 1 & 4 \\
44 & 0 & 5 & 0 & 0 & 1 & 1 & 3 \\
44 & 1 & 4 & 1 & 1 & 1 & 2 & 3 \\
44 & 2 & 4 & 0 & 0 & 1 & 1 & 3 \\
44 & 3 & 6 & 1 & 3 & 1 & 1 & 4 \\
\bottomrule\end{tabular}\end{center}\end{table}
\begin{table}[ht]\caption{DeepSeek \texttt{gepa\_llm} solved counts by family and round. \texttt{incident\_track}, \texttt{postmortem}, and \texttt{onboarding} are omitted (0 on every pass).}\label{tab:famgepa_llm}\begin{center}\scriptsize\setlength{\tabcolsep}{3pt}
\begin{tabular}{rrrrrrrr}\toprule Seed & Round & \rotatebox{65}{\texttt{catalog\_discover}} & \rotatebox{65}{\texttt{brief\_report}} & \rotatebox{65}{\texttt{checklist}} & \rotatebox{65}{\texttt{csv\_audit}} & \rotatebox{65}{\texttt{access\_denied}} & \rotatebox{65}{\texttt{do\_not\_publish}} \\\midrule
42 & 0 & 5 & 1 & 1 & 1 & 1 & 5 \\
42 & 1 & 5 & 0 & 1 & 1 & 1 & 2 \\
42 & 2 & 5 & 1 & 1 & 1 & 2 & 6 \\
42 & 3 & 6 & 2 & 2 & 2 & 1 & 4 \\
43 & 0 & 5 & 0 & 1 & 1 & 1 & 4 \\
43 & 1 & 5 & 1 & 1 & 1 & 1 & 3 \\
43 & 2 & 5 & 2 & 1 & 2 & 2 & 4 \\
43 & 3 & 4 & 2 & 0 & 1 & 1 & 4 \\
44 & 0 & 3 & 2 & 0 & 1 & 1 & 3 \\
44 & 1 & 5 & 2 & 0 & 1 & 2 & 4 \\
44 & 2 & 5 & 2 & 0 & 1 & 1 & 5 \\
44 & 3 & 7 & 2 & 1 & 1 & 2 & 4 \\
\bottomrule\end{tabular}\end{center}\end{table}
\begin{table}[ht]\caption{DeepSeek \texttt{ace\_llm} solved counts by family and round. \texttt{incident\_track}, \texttt{postmortem}, and \texttt{onboarding} are omitted (0 on every pass).}\label{tab:famace_llm}\begin{center}\scriptsize\setlength{\tabcolsep}{3pt}
\begin{tabular}{rrrrrrrr}\toprule Seed & Round & \rotatebox{65}{\texttt{catalog\_discover}} & \rotatebox{65}{\texttt{brief\_report}} & \rotatebox{65}{\texttt{checklist}} & \rotatebox{65}{\texttt{csv\_audit}} & \rotatebox{65}{\texttt{access\_denied}} & \rotatebox{65}{\texttt{do\_not\_publish}} \\\midrule
42 & 0 & 5 & 0 & 2 & 1 & 1 & 4 \\
42 & 1 & 5 & 1 & 0 & 1 & 1 & 4 \\
42 & 2 & 6 & 2 & 0 & 2 & 2 & 3 \\
42 & 3 & 4 & 1 & 2 & 2 & 1 & 5 \\
43 & 0 & 5 & 2 & 0 & 1 & 1 & 4 \\
43 & 1 & 4 & 0 & 1 & 1 & 1 & 3 \\
43 & 2 & 5 & 2 & 0 & 1 & 1 & 4 \\
43 & 3 & 7 & 2 & 3 & 2 & 1 & 4 \\
44 & 0 & 5 & 1 & 0 & 1 & 2 & 3 \\
44 & 1 & 6 & 0 & 0 & 1 & 1 & 3 \\
44 & 2 & 3 & 1 & 2 & 1 & 1 & 5 \\
44 & 3 & 5 & 2 & 0 & 1 & 1 & 4 \\
\bottomrule\end{tabular}\end{center}\end{table}
\begin{table}[ht]\caption{DeepSeek \texttt{expel\_llm} solved counts by family and round. \texttt{incident\_track}, \texttt{postmortem}, and \texttt{onboarding} are omitted (0 on every pass).}\label{tab:famexpel_llm}\begin{center}\scriptsize\setlength{\tabcolsep}{3pt}
\begin{tabular}{rrrrrrrr}\toprule Seed & Round & \rotatebox{65}{\texttt{catalog\_discover}} & \rotatebox{65}{\texttt{brief\_report}} & \rotatebox{65}{\texttt{checklist}} & \rotatebox{65}{\texttt{csv\_audit}} & \rotatebox{65}{\texttt{access\_denied}} & \rotatebox{65}{\texttt{do\_not\_publish}} \\\midrule
42 & 0 & 5 & 0 & 0 & 1 & 1 & 3 \\
42 & 1 & 5 & 2 & 0 & 1 & 1 & 4 \\
42 & 2 & 4 & 1 & 1 & 1 & 1 & 5 \\
42 & 3 & 7 & 3 & 1 & 1 & 1 & 4 \\
43 & 0 & 5 & 1 & 0 & 1 & 2 & 4 \\
43 & 1 & 5 & 1 & 0 & 0 & 2 & 4 \\
43 & 2 & 3 & 0 & 0 & 1 & 1 & 3 \\
43 & 3 & 6 & 2 & 2 & 1 & 1 & 5 \\
44 & 0 & 5 & 1 & 1 & 1 & 1 & 4 \\
44 & 1 & 5 & 1 & 0 & 1 & 1 & 3 \\
44 & 2 & 4 & 1 & 1 & 1 & 1 & 3 \\
44 & 3 & 7 & 2 & 4 & 1 & 3 & 3 \\
\bottomrule\end{tabular}\end{center}\end{table}
\section{Seed-42 family counts outside DeepSeek}
\label{app:transferfam}
DeepSeek family counts are Table~\ref{tab:v1fam} and Appendix~\ref{app:famcount}. The cells below are the seed-42 transfer. Entries are solved counts at round 0 and at the last recorded round.
\begin{longtable}{llrr}
\caption{GLM 5.3 and MiniMax M2.5 solved counts, round 0 to last round.}\label{tab:transferfam}\\
\toprule Cell & Family & Round 0 & Last \\
\midrule\endfirsthead
\toprule Cell & Family & Round 0 & Last \\
\midrule\endhead
\bottomrule\endfoot
GLM no\_evolve & access\_denied & 2/6 & 3/6 \\
GLM no\_evolve & brief\_report & 3/5 & 1/5 \\
GLM no\_evolve & catalog\_discover & 4/8 & 5/8 \\
GLM no\_evolve & checklist & 2/5 & 0/5 \\
GLM no\_evolve & csv\_audit & 1/4 & 1/4 \\
GLM no\_evolve & do\_not\_publish & 5/6 & 4/6 \\
GLM gepa & access\_denied & 2/6 & 2/6 \\
GLM gepa & brief\_report & 3/5 & 2/5 \\
GLM gepa & catalog\_discover & 6/8 & 2/8 \\
GLM gepa & checklist & 3/5 & 2/5 \\
GLM gepa & csv\_audit & 1/4 & 1/4 \\
GLM gepa & do\_not\_publish & 4/6 & 4/6 \\
GLM ace & access\_denied & 2/6 & 1/6 \\
GLM ace & brief\_report & 3/5 & 2/5 \\
GLM ace & catalog\_discover & 6/8 & 6/8 \\
GLM ace & checklist & 2/5 & 1/5 \\
GLM ace & csv\_audit & 1/4 & 1/4 \\
GLM ace & do\_not\_publish & 5/6 & 4/6 \\
GLM expel & access\_denied & 1/6 & 2/6 \\
GLM expel & brief\_report & 3/5 & 2/5 \\
GLM expel & catalog\_discover & 3/8 & 6/8 \\
GLM expel & checklist & 1/5 & 0/5 \\
GLM expel & csv\_audit & 2/4 & 1/4 \\
GLM expel & do\_not\_publish & 5/6 & 4/6 \\
MM no\_evolve & access\_denied & 1/6 & 1/6 \\
MM no\_evolve & brief\_report & 0/5 & 1/5 \\
MM no\_evolve & catalog\_discover & 5/8 & 4/8 \\
MM no\_evolve & checklist & 1/5 & 0/5 \\
MM no\_evolve & do\_not\_publish & 1/6 & 3/6 \\
MM gepa & access\_denied & 1/6 & 1/6 \\
MM gepa & brief\_report & 1/5 & 1/5 \\
MM gepa & catalog\_discover & 2/8 & 3/8 \\
MM gepa & checklist & 2/5 & 1/5 \\
MM gepa & do\_not\_publish & 1/6 & 3/6 \\
MM ace & access\_denied & 1/6 & 1/6 \\
MM ace & brief\_report & 1/5 & 1/5 \\
MM ace & catalog\_discover & 5/8 & 4/8 \\
MM ace & checklist & 2/5 & 1/5 \\
MM ace & csv\_audit & 0/4 & 1/4 \\
MM ace & do\_not\_publish & 4/6 & 3/6 \\
MM expel & access\_denied & 1/6 & 3/6 \\
MM expel & brief\_report & 1/5 & 1/5 \\
MM expel & catalog\_discover & 4/8 & 4/8 \\
MM expel & checklist & 1/5 & 2/5 \\
MM expel & csv\_audit & 0/4 & 1/4 \\
MM expel & do\_not\_publish & 2/6 & 2/6 \\
\end{longtable}
Rows that stay at zero solved tasks on both ends are omitted. \texttt{incident\_track}, \texttt{postmortem}, and \texttt{onboarding} are zero on every round of every seed-42 cell, matching the DeepSeek passes.
\section{Failure reasons outside the DeepSeek office cells}
\label{app:failmore}
Table~\ref{tab:v1fail} is DeepSeek office-v1. Table~\ref{tab:failmore} is the seed-42 transfer on the same pack. Counts are \texttt{max\_steps} / \texttt{verifier\_fail}. A dash is a round the runner did not start.
\begin{longtable}{llrrrr}
\caption{Seed-42 office-v1 transfer. Counts are \texttt{max\_steps} / \texttt{verifier\_fail}.}\label{tab:failmore}\\
\toprule Cell & Seed & r0 & r1 & r2 & r3 \\
\midrule\endfirsthead
\toprule Cell & Seed & r0 & r1 & r2 & r3 \\
\midrule\endhead
\bottomrule\endfoot
GLM no\_evolve & 42 & 17/6 & 22/4 & 14/5 & 22/4 \\
GLM gepa & 42 & 14/7 & 12/10 & 20/5 & 22/5 \\
GLM ace & 42 & 18/3 & 16/5 & 21/5 & 20/5 \\
GLM expel & 42 & 20/5 & 18/2 & 23/3 & 23/2 \\
MM no\_evolve & 42 & 27/5 & 29/1 & 30/2 & 28/3 \\
MM gepa & 42 & 31/2 & 29/3 & 25/4 & 29/2 \\
MM ace & 42 & 23/4 & 29/3 & 29/0 & --- \\
MM expel & 42 & 28/3 & 28/3 & 31/1 & 27/0 \\
\end{longtable}
Every recorded failure in Table~\ref{tab:failmore} is \texttt{max\_steps} or \texttt{verifier\_fail}.

\begin{table}[H]
\caption{Manufacturing-v1 \texttt{no\_evolve}. Counts are \texttt{below\_threshold} / \texttt{max\_steps}. These passes have no \texttt{verifier\_fail}.}
\label{tab:mfgfail}
\begin{center}
\scriptsize
\begin{tabular}{rrrrr}
\toprule
Seed & r0 & r1 & r2 & r3 \\
\midrule
42 & 17/9 & 15/8 & 13/10 & 16/9 \\
43 & 16/8 & 17/7 & 14/8 & 15/8 \\
44 & 15/9 & 16/8 & 15/9 & 15/8 \\
\bottomrule
\end{tabular}
\end{center}
\end{table}
\section{Eligibility flags}
\label{app:flags}
\texttt{gated} is false when final held-out regression exceeds 0.20, the run is over the evolution-token cap, the trace origin is not self, the anchor flag is not ok, the artifact does not match, contamination is set, $M_0$ is unbound, or the gate is off. \texttt{paper\_min} repeats those gated reasons and also fails a pack that is under 40/10/20, has no OOD slice, overlaps split ids, or leaves the default mix and schedule. Coding has no OOD slice, so this runner reason fires there; the paper still reports that board and states the single family in Section~\ref{sec:coding}. On these office-v1 cells the pack size is 80/20/40, so a \texttt{paper\_min} mark that travels with \texttt{gated} is the copied regression or budget reason, not a short pack. \texttt{main\_table\_eligible} can still be true for a frozen \texttt{no\_evolve} cell whose regression is at most 0.20. That flag does not mean the policy changed. The Main column is the runner's office-lane flag. It is not this paper's ranking; coding is the primary scoreboard (Section~\ref{sec:coding}).
\begin{longtable}{llrl}
\caption{Stored eligibility reasons for the 20 office-v1 cells. Main is the runner's office-lane flag, not the coding scoreboard.}\label{tab:flags}\\
\toprule Cell & Seed & Main & Reasons \\
\midrule\endfirsthead
\toprule Cell & Seed & Main & Reasons \\
\midrule\endhead
\bottomrule\endfoot
DS ace & 42 & false & gated, paper\_min \\
DS ace & 43 & true & eligible \\
DS ace & 44 & false & gated, paper\_min \\
DS expel & 42 & true & eligible \\
DS expel & 43 & true & eligible \\
DS expel & 44 & true & eligible \\
DS gepa & 42 & false & gated, paper\_min \\
DS gepa & 43 & false & gated, paper\_min \\
DS gepa & 44 & true & eligible \\
DS no\_evolve & 42 & true & eligible \\
DS no\_evolve & 43 & false & gated, paper\_min \\
DS no\_evolve & 44 & true & eligible \\
GLM ace & 42 & false & gated, paper\_min, over\_budget \\
GLM expel & 42 & false & gated, paper\_min, over\_budget \\
GLM gepa & 42 & false & gated, paper\_min, over\_budget \\
GLM no\_evolve & 42 & false & gated, paper\_min \\
MM ace & 42 & false & gated, paper\_min, over\_budget \\
MM expel & 42 & true & eligible \\
MM gepa & 42 & false & gated, paper\_min \\
MM no\_evolve & 42 & false & gated, paper\_min \\
\end{longtable}
\section{Office grids}
\label{sec:officev1}

\paragraph{Setup.}
One hosted model is both the tool actor and the evolver. The actor is the TEB Standard Tool Actor: temperature 0, one user message, neutral prompt, at most 20 steps. Budget class is B0. The pack is frozen \texttt{office-v1}: catalog-workspace jobs, deterministic checks, no gold fields in the observation (all three packs: Appendix~\ref{app:data}). Evolution chunks are 27/27/26 of the 80 evolution tasks. Seeds $\{42,43,44\}$ shuffle evolution order, not held-out membership. The three-seed methods were fixed from a development screen on validation gain versus model-matched \texttt{no\_evolve}, then lower validation regression, then lower tokens, before formal held-out output was read: \texttt{gepa\_llm}, \texttt{ace\_llm}~\citep{zhang2026ace}, \texttt{expel\_llm}. Their mean screen gains were $0$, $-5.6$, and $-11.1$ pp (Appendix~\ref{app:screen}). \texttt{awm\_llm} and \texttt{reasoning\_bank}~\citep{ouyang2026reasoningbank} failed the screen's artifact check. \texttt{dgm} and \texttt{clin\_llm}~\citep{majumder2024clin} ranked below and were not given this matrix. Every finished cell has \texttt{contamination=false} and \texttt{adapter\_honesty=ok} (a labeled text update, not the upstream implementation).

\paragraph{What a pass is.}
The actor's tools are \texttt{list\_files}, \texttt{read\_file}, \texttt{write\_file}, \texttt{publish\_artifact}, \texttt{grant\_access}, and \texttt{submit}. A failed \texttt{submit} returns the failed check names and leaves the episode open. Success is every terminal check passing. Gold files are not in the observation~\citep{elangovan2021leakage}. Four grade classes add checks on top of that contract: \texttt{discover} (names and list only), \texttt{extract} (bound phrases, length cap), \texttt{report} (facts under the required heading), and \texttt{structured} (strict JSON, no extra keys). Eight of the forty held-out tasks are \texttt{structured}. They stay inside the same Success denominator. Office Success is instruction following and rule-bounded delivery~\citep{zhou2023ifeval,ribeiro2020checklist}. It is not a writing-quality score.

\subsection{Frozen band}
Table~\ref{tab:v1noise} and the first panel of Figure~\ref{fig:v1rounds} are the control. DeepSeek v4 Pro and GLM 5.3 each span 17.5 pp with no context update. MiniMax M2.5~\citep{minimax2026m2} spans 5.0 pp. DeepSeek seed 42's path is $35.0, 32.5, 27.5, 45.0$, all \texttt{agent\_version}=$M_0$. Actor, prompt, and held-out ids are identical across these passes; the hosted endpoint is not deterministic at temperature 0~\citep{atil2024nondeterminism}, and seeds do not control it. Round 0 itself is $35.0$, $27.5$, and $25.0$ across the three seeds, so Gain's baseline is one draw from this band. Seed 43's regression is $0.231$, so the runner marks that frozen cell ineligible: the regression gate fires on rescore noise. On seed 42, every frozen MiniMax pass (20--25\%) is below every frozen GLM pass (35--52.5\%), so the pack separates models even where it cannot resolve updates.

\begin{table}[ht]
\caption{Frozen-policy band. Range is max minus min of the four held-out passes, in percentage points.}
\label{tab:v1noise}
\begin{center}
\scriptsize
\setlength{\tabcolsep}{4pt}
\begin{tabular}{llrrrrr}
\toprule
Model & Seed & r0 & r1 & r2 & r3 & Range \\
\midrule
DeepSeek v4 Pro & 42 & 35.0 & 32.5 & 27.5 & 45.0 & 17.5 \\
DeepSeek v4 Pro & 43 & 27.5 & 35.0 & 32.5 & 35.0 & 7.5 \\
DeepSeek v4 Pro & 44 & 25.0 & 30.0 & 22.5 & 40.0 & 17.5 \\
GLM 5.3 & 42 & 42.5 & 35.0 & 52.5 & 35.0 & 17.5 \\
MiniMax M2.5 & 42 & 20.0 & 25.0 & 20.0 & 22.5 & 5.0 \\
\bottomrule
\end{tabular}
\end{center}
\end{table}

\subsection{Paired excess}
Figure~\ref{fig:v1rounds} shows that \texttt{no\_evolve} moves with the adapters. Figure~\ref{fig:v1excess} shades the widest DeepSeek band, $\pm 17.5$ pp, which is wider than seed 43's own $7.5$ pp band. Means and the per-seed reading are in Section~\ref{sec:officectr}. Table~\ref{tab:v1excess} is the cell record. Its Gain column still contains the frozen movement in Table~\ref{tab:v1noise}.

\begin{figure}[ht]
\centering
\includegraphics[width=\linewidth]{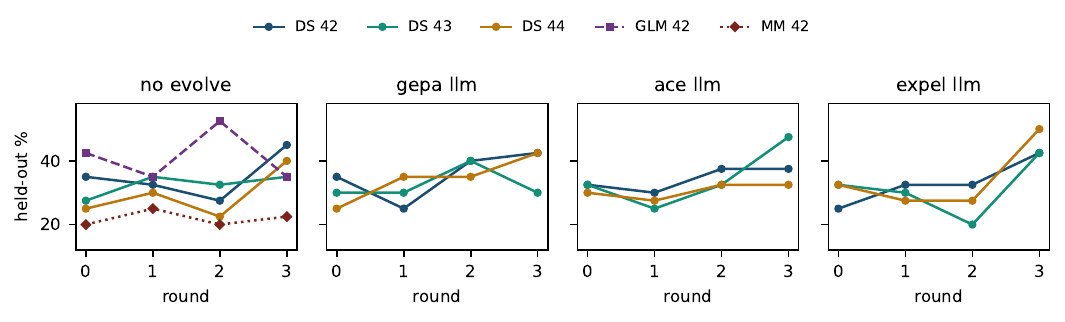}
\caption{Held-out success by round. The first panel is frozen \texttt{no\_evolve}, including GLM 5.3 and MiniMax M2.5 at seed 42. DeepSeek seed 42 goes from 27.5\% at round 2 to 45\% at round 3 with no context update. The other panels are DeepSeek only. \texttt{no\_evolve} moves as much as the adapters.}
\label{fig:v1rounds}
\end{figure}

\begin{figure}[ht]
\centering
\includegraphics[width=\linewidth]{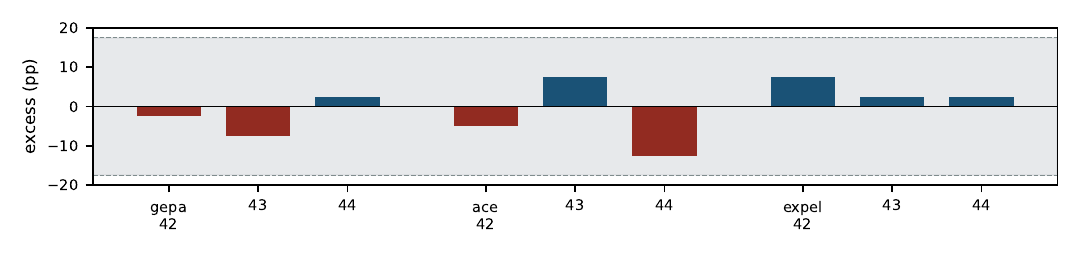}
\caption{DeepSeek office paired excess. Shading is the widest seed band, $\pm 17.5$ pp, not seed 43's own $7.5$ pp band. Each seed's range is the Range column of Table~\ref{tab:v1noise}.}
\label{fig:v1excess}
\end{figure}

\subsection{Same agent, different score}
A version change is an \texttt{evolve()} accept, including 0 pp ties under \texttt{min\_delta}${=}0$: DeepSeek \texttt{ace\_llm} seed 42 ($35{\to}35$ to $M_3$), \texttt{expel\_llm} seed 44 ($15{\to}15$, $30{\to}30$), and MiniMax \texttt{gepa\_llm} (three ties to $M_3$; Appendix~\ref{app:gatesv1}). A score change without a version change is a rescore~\citep{atil2024nondeterminism}. \texttt{expel\_llm} seed 43 is $M_2$ at both round 2 (20.0\%) and round 3 (42.5\%): $+22.5$ pp on one artifact. \texttt{gepa\_llm} seed 43 rescores $M_1$ from 40.0\% to 30.0\%. GLM \texttt{ace\_llm} and \texttt{expel\_llm} stay $M_0$: rounds 1 and 2 roll back, and round 3 hits the evolution-token cap. Their held-out paths still span 12.5 pp and 15.0 pp. Table~\ref{tab:v1jump} lists same-version gaps of at least 10 pp. Table~\ref{tab:v1ver} is the accepted version after every round. Figure~\ref{fig:v1heat} is every finished cell, next to the full grid. MiniMax \texttt{ace\_llm} has no round 3.

\begin{table}[ht]
\caption{Accepted \texttt{agent\_version} after each office-v1 round. A repeated version means the artifact did not change. On \texttt{no\_evolve} that is every round.}
\label{tab:v1ver}
\begin{center}
\scriptsize
\setlength{\tabcolsep}{3.5pt}
\begin{tabular}{llrllll}
\toprule
Model & Method & Seed & r0 & r1 & r2 & r3 \\
\midrule
DS & no\_evolve & 42 & $M_0$ & $M_0$ & $M_0$ & $M_0$ \\
DS & no\_evolve & 43 & $M_0$ & $M_0$ & $M_0$ & $M_0$ \\
DS & no\_evolve & 44 & $M_0$ & $M_0$ & $M_0$ & $M_0$ \\
DS & gepa & 42 & $M_0$ & $M_0$ & $M_1$ & $M_1$ \\
DS & gepa & 43 & $M_0$ & $M_0$ & $M_1$ & $M_1$ \\
DS & gepa & 44 & $M_0$ & $M_0$ & $M_0$ & $M_1$ \\
DS & ace & 42 & $M_0$ & $M_1$ & $M_2$ & $M_3$ \\
DS & ace & 43 & $M_0$ & $M_0$ & $M_0$ & $M_1$ \\
DS & ace & 44 & $M_0$ & $M_1$ & $M_2$ & $M_2$ \\
DS & expel & 42 & $M_0$ & $M_1$ & $M_2$ & $M_3$ \\
DS & expel & 43 & $M_0$ & $M_1$ & $M_2$ & $M_2$ \\
DS & expel & 44 & $M_0$ & $M_1$ & $M_2$ & $M_3$ \\
GLM & no\_evolve & 42 & $M_0$ & $M_0$ & $M_0$ & $M_0$ \\
GLM & gepa & 42 & $M_0$ & $M_1$ & $M_1$ & $M_1$ \\
GLM & ace & 42 & $M_0$ & $M_0$ & $M_0$ & $M_0$ \\
GLM & expel & 42 & $M_0$ & $M_0$ & $M_0$ & $M_0$ \\
MM & no\_evolve & 42 & $M_0$ & $M_0$ & $M_0$ & $M_0$ \\
MM & gepa & 42 & $M_0$ & $M_1$ & $M_2$ & $M_3$ \\
MM & ace & 42 & $M_0$ & $M_1$ & $M_1$ & --- \\
MM & expel & 42 & $M_0$ & $M_0$ & $M_0$ & $M_1$ \\
\bottomrule
\end{tabular}
\end{center}
\end{table}

\begin{table}[ht]
\caption{Consecutive held-out passes with the same \texttt{agent\_version} and an absolute gap of at least 10 pp. The accepted artifact did not change between the two scores.}
\label{tab:v1jump}
\begin{center}
\scriptsize
\setlength{\tabcolsep}{3pt}
\begin{tabular}{llrrl}
\toprule
Cell & Version & From & To & Gap \\
\midrule
DS no\_evolve 42 & $M_0$ & 27.5 & 45.0 & 17.5 \\
DS no\_evolve 44 & $M_0$ & 22.5 & 40.0 & 17.5 \\
DS gepa 42 & $M_0$ & 35.0 & 25.0 & 10.0 \\
DS gepa 43 & $M_1$ & 40.0 & 30.0 & 10.0 \\
DS gepa 44 & $M_0$ & 25.0 & 35.0 & 10.0 \\
DS expel 43 & $M_2$ & 20.0 & 42.5 & 22.5 \\
GLM no\_evolve 42 & $M_0$ & 35.0 & 52.5 & 17.5 \\
GLM no\_evolve 42 & $M_0$ & 52.5 & 35.0 & 17.5 \\
GLM ace 42 & $M_0$ & 47.5 & 35.0 & 12.5 \\
GLM expel 42 & $M_0$ & 37.5 & 50.0 & 12.5 \\
GLM expel 42 & $M_0$ & 50.0 & 35.0 & 15.0 \\
\bottomrule
\end{tabular}
\end{center}
\end{table}

\subsection{Full grid}
Table~\ref{tab:v1grid} reports all 20 office-v1 cells; Figure~\ref{fig:v1heat} shows their round-wise scores.

\begin{figure}[ht]
\centering
\includegraphics[width=0.58\linewidth]{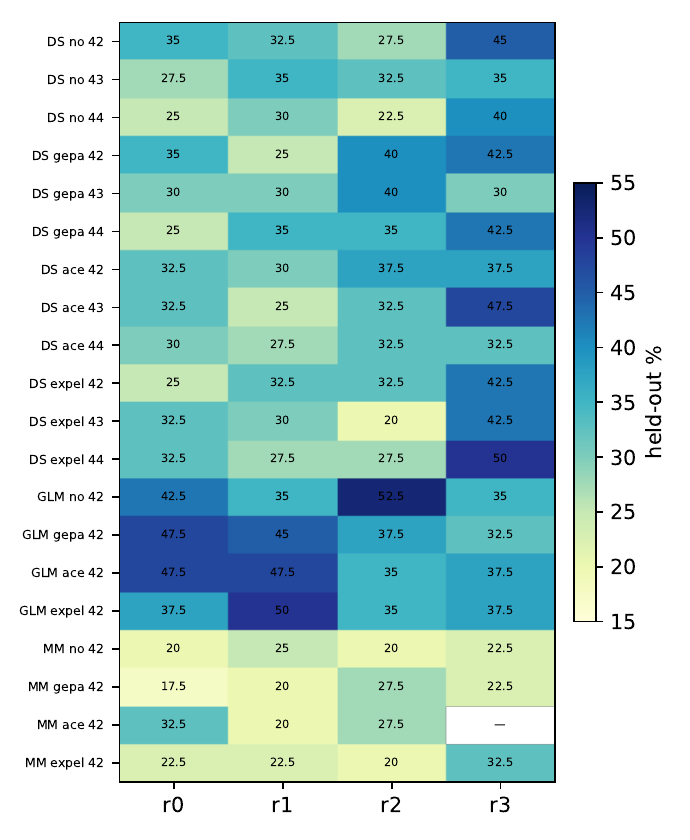}
\caption{Held-out success (\%) at each round for all 20 cells. A dash is a round the runner did not start (MiniMax \texttt{ace\_llm} round 3).}
\label{fig:v1heat}
\end{figure}
 Flags: reg.\ means final regression $>0.20$; budget means \texttt{over\_budget}. GLM rows that show a round-3 success still recorded that held-out pass and were then marked over the evolution-token cap. GLM \texttt{gepa\_llm} regression is $0.200$, which does not by itself exceed $0.20$; the cell is ineligible because round 3 crossed $500{,}000$ evolution tokens. DeepSeek \texttt{expel\_llm} seeds 42 and 44 reach $M_3$. Their excesses are $+7.5$ and $+2.5$ pp. Eligible DeepSeek excesses do not exceed the seed's own range in Table~\ref{tab:v1noise}. \texttt{ace\_llm} seed 43 is $+7.5$ pp, equal to that seed's $7.5$ pp range. The other positive eligible excesses are \texttt{expel\_llm} $+7.5,+2.5,+2.5$ and \texttt{gepa\_llm} seed 44 $+2.5$, against the $17.5$ pp ranges of seeds 42 and 44 (Appendix~\ref{app:flags}).

\begin{table}[ht]
\caption{All 20 office-v1 cells. Success columns are held-out percent. Gain and excess are percentage points. Excess is blank on \texttt{no\_evolve}. DS / GLM / MM are DeepSeek v4 Pro, GLM 5.3, and MiniMax M2.5.}
\label{tab:v1grid}
\begin{center}
\footnotesize
\setlength{\tabcolsep}{3pt}
\begin{tabular}{llrrrrrrrl}
\toprule
Model & Method & Seed & r0 & r1 & r2 & r3 & Gain & Excess & Flag \\
\midrule
DS & no\_evolve & 42 & 35.0 & 32.5 & 27.5 & 45.0 & $+10.0$ & --- & \\
DS & no\_evolve & 43 & 27.5 & 35.0 & 32.5 & 35.0 & $+7.5$ & --- & reg. \\
DS & no\_evolve & 44 & 25.0 & 30.0 & 22.5 & 40.0 & $+15.0$ & --- & \\
DS & gepa & 42 & 35.0 & 25.0 & 40.0 & 42.5 & $+7.5$ & $-2.5$ & reg. \\
DS & gepa & 43 & 30.0 & 30.0 & 40.0 & 30.0 & $0.0$ & $-7.5$ & reg. \\
DS & gepa & 44 & 25.0 & 35.0 & 35.0 & 42.5 & $+17.5$ & $+2.5$ & \\
DS & ace & 42 & 32.5 & 30.0 & 37.5 & 37.5 & $+5.0$ & $-5.0$ & reg. \\
DS & ace & 43 & 32.5 & 25.0 & 32.5 & 47.5 & $+15.0$ & $+7.5$ & \\
DS & ace & 44 & 30.0 & 27.5 & 32.5 & 32.5 & $+2.5$ & $-12.5$ & reg. \\
DS & expel & 42 & 25.0 & 32.5 & 32.5 & 42.5 & $+17.5$ & $+7.5$ & \\
DS & expel & 43 & 32.5 & 30.0 & 20.0 & 42.5 & $+10.0$ & $+2.5$ & \\
DS & expel & 44 & 32.5 & 27.5 & 27.5 & 50.0 & $+17.5$ & $+2.5$ & \\
GLM & no\_evolve & 42 & 42.5 & 35.0 & 52.5 & 35.0 & $-7.5$ & --- & reg. \\
GLM & gepa & 42 & 47.5 & 45.0 & 37.5 & 32.5 & $-15.0$ & $-7.5$ & budget \\
GLM & ace & 42 & 47.5 & 47.5 & 35.0 & 37.5 & $-10.0$ & $-2.5$ & budget \\
GLM & expel & 42 & 37.5 & 50.0 & 35.0 & 37.5 & $0.0$ & $+7.5$ & budget \\
MM & no\_evolve & 42 & 20.0 & 25.0 & 20.0 & 22.5 & $+2.5$ & --- & reg. \\
MM & gepa & 42 & 17.5 & 20.0 & 27.5 & 22.5 & $+5.0$ & $+2.5$ & reg. \\
MM & ace & 42 & 32.5 & 20.0 & 27.5 & --- & $-5.0$ & $-7.5$ & budget \\
MM & expel & 42 & 22.5 & 22.5 & 20.0 & 32.5 & $+10.0$ & $+7.5$ & \\
\bottomrule
\end{tabular}
\end{center}
\end{table}

\subsection{Seed-42 transfer}
MiniMax M2.5 and GLM 5.3 are one seed. The reading is in Section~\ref{sec:officectr}. Figure~\ref{fig:v1exp} and Table~\ref{tab:v1transfer} are the cell record.

\begin{figure}[ht]
\centering
\includegraphics[width=\linewidth]{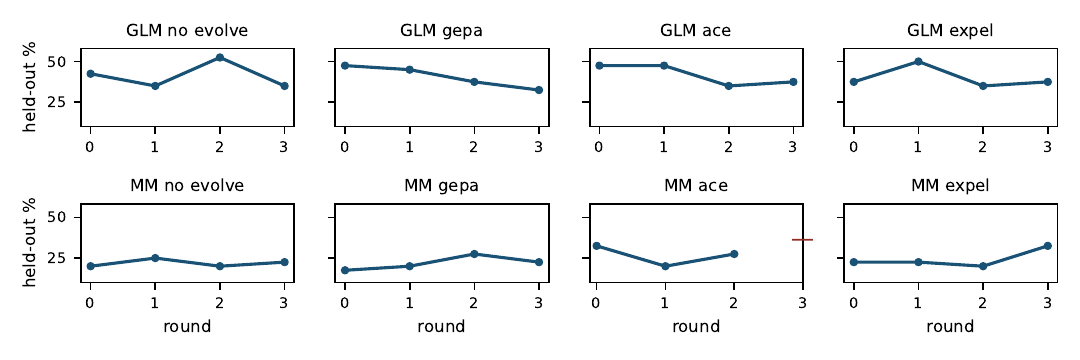}
\caption{Seed-42 transfer. Top row: GLM 5.3. Bottom row: MiniMax M2.5. A dash is a round the runner did not start (MiniMax \texttt{ace\_llm} round 3). These curves are not a second three-seed confirmation.}
\label{fig:v1exp}
\end{figure}

\begin{table}[ht]
\caption{Seed-42 transfer. Excess is against that model's \texttt{no\_evolve}. Not ranked with DeepSeek.}
\label{tab:v1transfer}
\begin{center}
\scriptsize
\setlength{\tabcolsep}{3pt}
\begin{tabular}{llrrrl}
\toprule
Model & Method & Succ. & Gain & Excess & Note \\
\midrule
MiniMax & no\_evolve & 22.5 & $+2.5$ & --- & reg.\ 0.25 \\
MiniMax & gepa & 22.5 & $+5.0$ & $+2.5$ & inside 5.0; reg.\ 0.273 \\
MiniMax & ace & 27.5 & $-5.0$ & $-7.5$ & stopped at round 2 \\
MiniMax & expel & 32.5 & $+10.0$ & $+7.5$ & 2.5 pp past a 5.0 band \\
GLM & no\_evolve & 35.0 & $-7.5$ & --- & reg.\ 0.381 \\
GLM & gepa & 32.5 & $-15.0$ & $-7.5$ & budget; reg.\ 0.20 \\
GLM & ace & 37.5 & $-10.0$ & $-2.5$ & budget; version stays $M_0$ \\
GLM & expel & 37.5 & $0.0$ & $+7.5$ & budget; version stays $M_0$ \\
\bottomrule
\end{tabular}
\end{center}
\end{table}

\subsection{Where the 40 tasks move}
\label{sec:families}

Held-out $n{=}40$ splits into nine families (Table~\ref{tab:v1fam}). Eight of the 80 evolution tasks are \texttt{content\_draft}; this held-out fingerprint contains none, so Gain cannot show that family. \texttt{access\_denied} and \texttt{do\_not\_publish} are held-out only ($n{=}6$ each)~\citep{koh2021wilds}. The other seven are not marked OOD on this held-out fingerprint. Figure~\ref{fig:v1fam} is the round-0 to final change in solved count for every DeepSeek cell. \texttt{incident\_track} ($n{=}3$), \texttt{postmortem} ($n{=}2$), and \texttt{onboarding} ($n{=}1$) stay at zero solved tasks on every round of all twelve DeepSeek cells, including the three \texttt{expel\_llm} runs that reach a new \texttt{agent\_version}. The movement that produces raw Gain is among families the frozen policy already solves on some passes: \texttt{catalog\_discover}, \texttt{brief\_report}, \texttt{checklist}, \texttt{csv\_audit}, \texttt{access\_denied}, and \texttt{do\_not\_publish}.

On \texttt{no\_evolve} seed 42 the final pass gains two \texttt{brief\_report}, two \texttt{checklist}, one \texttt{catalog\_discover}, and one \texttt{csv\_audit}, and loses one \texttt{access\_denied} and one \texttt{do\_not\_publish}. \texttt{expel\_llm} seed 44, the highest final Success in the matrix (50\%), moves \texttt{checklist} from $1/5$ to $4/5$ and \texttt{catalog\_discover} from $5/8$ to $7/8$. Those are the same families that move when the agent string does not change. A family rate on $n{\le}8$ flips when one or two tasks flip. That is the 2.5 pp task grain inside the 17.5 pp band (Appendix~\ref{app:examples}).

\begin{table}[ht]
\caption{Office-v1 held-out families. $n$ sums to 40. OOD families do not appear in evolution.}
\label{tab:v1fam}
\begin{center}
\scriptsize
\setlength{\tabcolsep}{4pt}
\begin{tabular}{lrrl}
\toprule
Family & $n$ & OOD & What the check is \\
\midrule
\texttt{catalog\_discover} & 8 & 0 & List \texttt{/catalog}, write and publish \texttt{catalog\_index.md} \\
\texttt{brief\_report} & 5 & 0 & Bounded summary; invented owners, dates, or claims fail \\
\texttt{checklist} & 5 & 0 & Required sections only \\
\texttt{csv\_audit} & 4 & 0 & Recompute stated counts or clean a table \\
\texttt{incident\_track} & 3 & 0 & Timeline, risk, and triage; unknowns stay unknown \\
\texttt{postmortem} & 2 & 0 & Facts and actions split; blame fails \\
\texttt{onboarding} & 1 & 0 & Numbered first-run steps \\
\texttt{access\_denied} & 6 & 6 & Tools return \texttt{permission\_denied} until \texttt{grant\_access} \\
\texttt{do\_not\_publish} & 6 & 6 & Write the draft and submit without \texttt{publish\_artifact} \\
\bottomrule
\end{tabular}
\end{center}
\end{table}

\begin{figure}[ht]
\centering
\includegraphics[width=\linewidth]{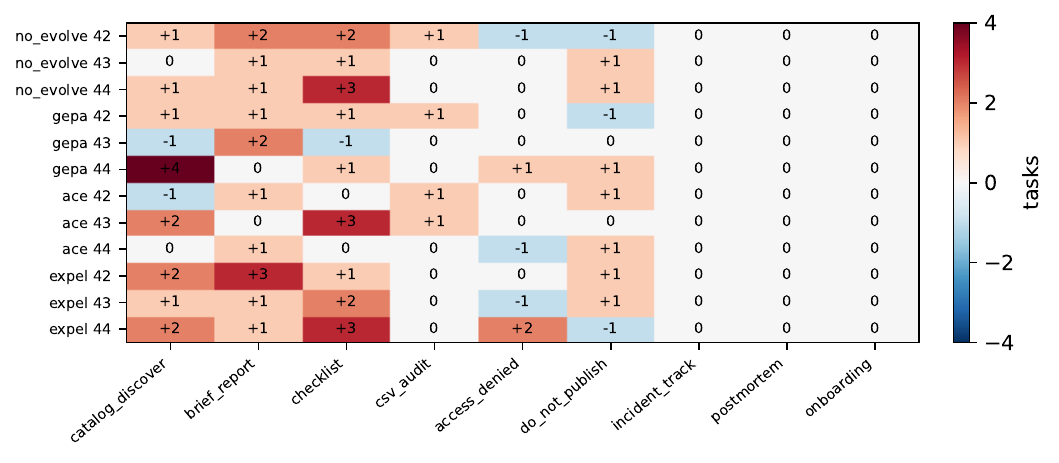}
\caption{DeepSeek family change from round 0 to the final held-out pass, in solved tasks. Zero-change families \texttt{incident\_track}, \texttt{postmortem}, and \texttt{onboarding} are included. Color is the task delta.}
\label{fig:v1fam}
\end{figure}

\subsection{Failure reasons}

\begin{table}[ht]
\caption{DeepSeek failure counts by round, \texttt{max\_steps} / \texttt{verifier\_fail}. The two counts sum to the number of held-out failures. No other failure reason appears.}
\label{tab:v1fail}
\begin{center}
\scriptsize
\setlength{\tabcolsep}{3pt}
\begin{tabular}{llrrrr}
\toprule
Method & Seed & r0 & r1 & r2 & r3 \\
\midrule
no\_evolve & 42 & 23/3 & 25/2 & 25/4 & 20/2 \\
no\_evolve & 43 & 24/5 & 21/5 & 24/3 & 25/1 \\
no\_evolve & 44 & 23/7 & 26/2 & 25/6 & 21/3 \\
gepa & 42 & 24/2 & 26/4 & 22/2 & 22/1 \\
gepa & 43 & 27/1 & 26/2 & 19/5 & 26/2 \\
gepa & 44 & 25/5 & 22/4 & 24/2 & 18/5 \\
ace & 42 & 21/6 & 24/4 & 21/4 & 22/3 \\
ace & 43 & 24/3 & 25/5 & 25/2 & 20/1 \\
ace & 44 & 23/5 & 25/4 & 26/1 & 22/5 \\
expel & 42 & 27/3 & 23/4 & 24/3 & 22/1 \\
expel & 43 & 24/3 & 24/4 & 24/8 & 20/3 \\
expel & 44 & 23/4 & 25/4 & 23/6 & 14/6 \\
\bottomrule
\end{tabular}
\end{center}
\end{table}

Raw Gain ranks \texttt{no\_evolve} seed 44 ($+15$ pp, $M_0$) with \texttt{gepa\_llm} seed 44 and two \texttt{expel\_llm} seeds at $+17.5$ pp. Paired excess ranks \texttt{gepa\_llm} seed 43 and \texttt{ace\_llm} seed 44 lowest ($-7.5$, $-12.5$). Table~\ref{tab:v1fail} is the movement: misses are the 20-step cap, and \texttt{no\_evolve} seed 42 goes $23,25,25,20$ \texttt{max\_steps} with no context update.

\section{Office artifact versions}
\label{app:officev1}

Table~\ref{tab:v1grid} in the main text is the score grid. Table~\ref{tab:v1ver} is the accepted artifact after each round. \texttt{no\_evolve} stays at $M_0$ on every seed. A repeated version means the gate rolled back or the runner rescored the same artifact. GLM \texttt{ace\_llm} and \texttt{expel\_llm} stay at $M_0$ because the evolution-token cap skipped \texttt{evolve()}. GLM \texttt{gepa\_llm} accepted $M_1$ and then rescored it. MiniMax \texttt{ace\_llm} has no round 3.

The evolution-token cap is 500{,}000 per round. After each evolution task the runner adds that task's token total, the sum of its steps. A step uses \texttt{usage.total\_tokens} when the server sends it, otherwise prompt tokens plus completion tokens. Validation tasks, held-out tasks, and model calls inside \texttt{evolve()} are not added. Crossing the cap skips \texttt{evolve()} for that round, marks the cell \texttt{over\_budget}, and skips later rounds when the violation is recorded before the next round.

Screen scores are not results. The screen ranked \texttt{gepa\_llm}, \texttt{ace\_llm}, and \texttt{expel\_llm} ahead of \texttt{dgm} and \texttt{clin\_llm} by mean validation gain against model-matched \texttt{no\_evolve}, then lower validation regression, then lower token cost. \texttt{awm\_llm} and \texttt{reasoning\_bank} failed the screen's artifact check. Those four methods were not given this held-out matrix. Running them on the same Gain definition would repeat the remeasurement in Table~\ref{tab:v1noise}.

Every finished cell reports \texttt{contamination=false}, \texttt{adapter\_honesty=ok}, \texttt{trace\_origin=self}, \texttt{trace\_mix=all}, \texttt{curriculum=none}, \texttt{budget\_class=B0}, \texttt{human\_edits=0}, and \texttt{m0\_source=measured}. The held-out fingerprint is identical across the DeepSeek cells that share the frozen pack. Temperature is 0. The client sends one user message. Inference timeout is 60 seconds, with retries on transport errors and HTTP 429/502/503/504. HTTP 402 is not retried. Held-out failures on the DeepSeek \texttt{no\_evolve} passes are \texttt{max\_steps} and verifier failures. A response cache of identical prompts would stabilize a rescore; it would not create the version changes in Table~\ref{tab:v1ver}.

\section{Manufacturing frozen repeat}
\label{app:mfg}

Manufacturing-v1 uses the same 80/20/40 shape, $R{=}3$, and DeepSeek v4 Pro \texttt{no\_evolve} on the Standard Tool Actor. The runner flags these cells \texttt{env\_not\_office}. This is the manufacturing frozen band, scored separately from coding. All three seeds stay at $M_0$, \texttt{contamination=false}, \texttt{adapter\_honesty=ok}, \texttt{over\_budget=false}. The four-pass ranges are 7.5, 5.0, and 2.5 pp. Failures on these passes are \texttt{below\_threshold} and \texttt{max\_steps}; \texttt{verifier\_fail} does not appear (Table~\ref{tab:mfgfail}). These ranges are not the office DeepSeek band and not the coding band (Appendix~\ref{app:coding}). Office allows 20 steps per task; this repeat allows 12.

\begin{table}[ht]
\caption{Manufacturing-v1, DeepSeek v4 Pro, \texttt{no\_evolve} only. Success is held-out percent. Version is $M_0$ at every round.}
\label{tab:mfg}
\begin{center}
\small
\begin{tabular}{rrrrrr}
\toprule
Seed & r0 & r1 & r2 & r3 & Range \\
\midrule
42 & 35.0 & 42.5 & 42.5 & 37.5 & 7.5 \\
43 & 40.0 & 40.0 & 45.0 & 42.5 & 5.0 \\
44 & 40.0 & 40.0 & 40.0 & 42.5 & 2.5 \\
\bottomrule
\end{tabular}
\end{center}
\end{table}

\section{Manufacturing family counts}
\label{app:mfgfam}
These are the finished DeepSeek v4 Pro \texttt{no\_evolve} cells in Table~\ref{tab:mfg}. Each entry is solved tasks on that family at rounds 0/1/2/3. \texttt{agent\_version} stays $M_0$. The office lane rejects the cells (\texttt{env\_not\_office}). Families with zero solved tasks on every pass are \texttt{accounts\_payable}, \texttt{advance\_payments}, \texttt{advance\_receipts}, \texttt{bpc\_report}, \texttt{receivable\_monthly}. \texttt{pc\_revenue} is zero at both ends; seed 44 round 1 solves one task.
\begin{longtable}{llccc}
\caption{Manufacturing-v1 solved counts by family at rounds 0--3, written r0/r1/r2/r3.}\label{tab:mfgfam}\\
\toprule Family & $n$ & Seed 42 & Seed 43 & Seed 44 \\
\midrule\endfirsthead
\toprule Family & $n$ & Seed 42 & Seed 43 & Seed 44 \\
\midrule\endhead
\bottomrule\endfoot
\texttt{accounts\_payable} & 3 & 0/0/0/0 & 0/0/0/0 & 0/0/0/0 \\
\texttt{advance\_payments} & 3 & 0/0/0/0 & 0/0/0/0 & 0/0/0/0 \\
\texttt{advance\_receipts} & 1 & 0/0/0/0 & 0/0/0/0 & 0/0/0/0 \\
\texttt{bpc\_report} & 7 & 0/0/0/0 & 0/0/0/0 & 0/0/0/0 \\
\texttt{commodity} & 3 & 1/2/2/2 & 2/2/2/2 & 2/2/2/2 \\
\texttt{electricity} & 3 & 3/3/3/3 & 3/3/3/3 & 3/3/3/3 \\
\texttt{expense\_detail} & 4 & 1/2/1/1 & 1/1/2/2 & 2/1/1/2 \\
\texttt{pc\_revenue} & 3 & 0/0/0/0 & 0/0/0/0 & 0/1/0/0 \\
\texttt{purchase\_receipt} & 3 & 3/3/3/3 & 3/3/3/3 & 3/3/3/3 \\
\texttt{receivable\_monthly} & 1 & 0/0/0/0 & 0/0/0/0 & 0/0/0/0 \\
\texttt{revenue\_cost} & 3 & 1/1/2/0 & 1/1/2/2 & 0/1/1/1 \\
\texttt{tax} & 3 & 2/3/3/3 & 3/3/3/3 & 3/3/3/3 \\
\texttt{utility} & 3 & 3/3/3/3 & 3/3/3/2 & 3/2/3/3 \\
\end{longtable}

\section{Coding frozen repeat}
\label{app:coding}

DeepSeek v4 Pro \texttt{no\_evolve} on the coding pack uses the same 80/20/40 shape and $R{=}3$. The runner flags these cells \texttt{env\_not\_office}. The four passes are Table~\ref{tab:coding}. All three seeds stay at $M_0$, \texttt{contamination=false}, \texttt{adapter\_honesty=ok}, \texttt{over\_budget=false}. The four-pass ranges are 7.5, 5.0, and 12.5 pp, with Gains of $-7.5$, $-2.5$, and $-10.0$ pp. The episode cap is 12 steps, not Office's 20. Every miss is that cap. Across the four passes the miss counts sum to 56, 60, and 63 on seeds 42, 43, and 44. Coding is a single family (\texttt{fix\_bug}, $n{=}40$), so there is no family split to tabulate. Final regression is $0.231$ and $0.308$ on seeds 42 and 44: the $0.20$ cutoff fires on frozen rescore noise here too, as on office seed 43.

\section{Extra office frozen passes}
\label{app:officeextra}

Six extra DeepSeek \texttt{no\_evolve} passes on the office-v1 held-out fingerprint (\texttt{afd748c6}, two per seed, $R{=}0$, no \texttt{evolve()} call) score 30.0 and 27.5 on seed 42, 27.5 and 22.5 on seed 43, and 30.0 and 27.5 on seed 44, all at $M_0$ with \texttt{contamination=false}. Merged with the original four passes, the per-seed ranges are 17.5, 12.5, and 17.5 pp. The widest stays 17.5 pp. Seed 43's own range widens from the 7.5 pp reported in Section~\ref{sec:officectr} to 12.5 pp, because one extra pass scores 22.5. These are the extra repeats the Limitations ask for.
\section{Coding grids}
\label{app:codinggrids}

These displays use the same 36 coding cells as Section~\ref{sec:coding}. Held-out is one family, so there is no family-delta figure. Every recorded miss is \texttt{max\_steps}; \texttt{verifier\_fail} does not appear.

\begin{table}[ht]
\caption{Coding frozen-policy passes for GLM and MiniMax. Gain and Range for all three models are Table~\ref{tab:codingfrozen}. DeepSeek passes are Table~\ref{tab:coding}. Range is max minus min of the four held-out passes, in percentage points.}
\label{tab:codingnoise}
\begin{center}
\scriptsize
\setlength{\tabcolsep}{4pt}
\begin{tabular}{llrrrrrr}
\toprule
Model & Seed & r0 & r1 & r2 & r3 & Gain & Range \\
\midrule
GLM & 42 & 72.5 & 72.5 & 60.0 & 70.0 & -2.5 & 12.5 \\
GLM & 43 & 60.0 & 65.0 & 67.5 & 67.5 & +7.5 & 7.5 \\
GLM & 44 & 67.5 & 70.0 & 75.0 & 60.0 & -7.5 & 15.0 \\
MiniMax & 42 & 30.0 & 37.5 & 40.0 & 35.0 & +5.0 & 10.0 \\
MiniMax & 43 & 37.5 & 35.0 & 37.5 & 37.5 & +0.0 & 2.5 \\
MiniMax & 44 & 40.0 & 37.5 & 27.5 & 40.0 & +0.0 & 12.5 \\
\bottomrule
\end{tabular}
\end{center}
\end{table}

\begin{figure}[ht]
\centering
\includegraphics[width=0.85\linewidth]{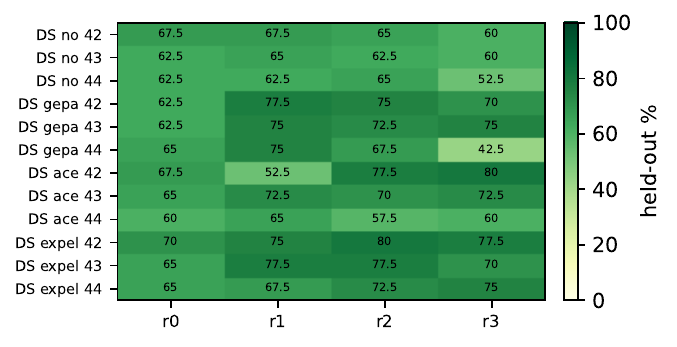}
\caption{DeepSeek coding held-out success (\%) at each round. Twelve cells: four methods by three seeds. Paired excess is Figure~\ref{fig:codingexcess}.}
\label{fig:codingheat}
\end{figure}

\begin{figure}[ht]
\centering
\includegraphics[width=0.85\linewidth]{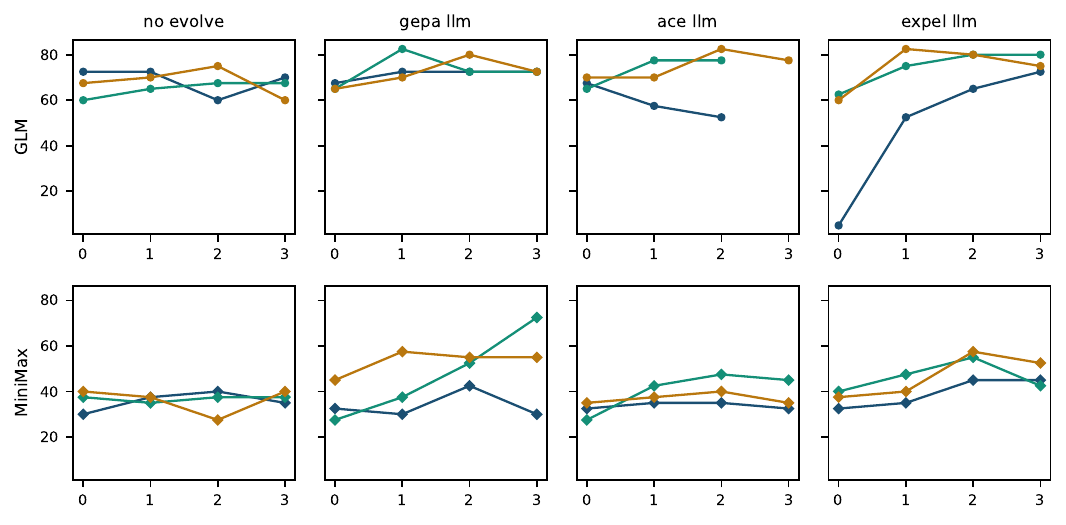}
\caption{GLM (top row) and MiniMax (bottom row) coding held-out success by round. Both models have seeds 42, 43, and 44. Line color is the seed.}
\label{fig:codingroundsother}
\end{figure}

\begin{figure}[ht]
\centering
\includegraphics[width=0.75\linewidth]{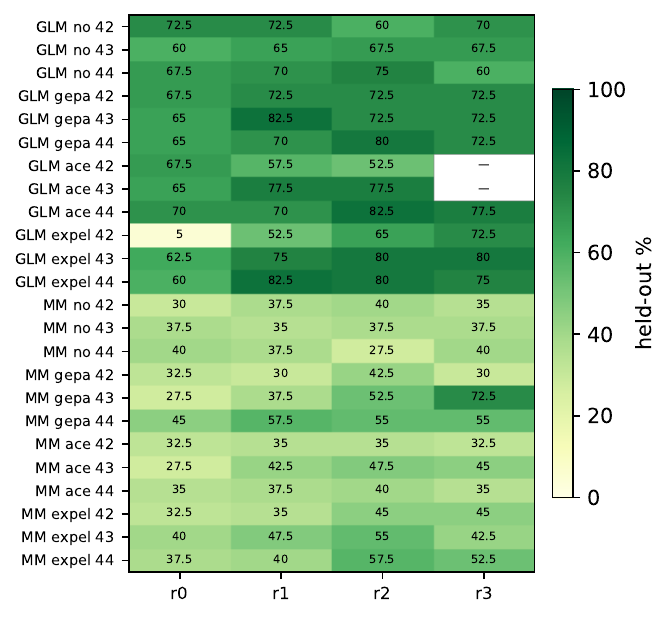}
\caption{GLM and MiniMax coding held-out success (\%) at each round. GLM \texttt{ace\_llm} seeds 42 and 43 have no round 3. Paired excess for these cells is Figure~\ref{fig:codingexcessother}.}
\label{fig:codingheatother}
\end{figure}

\begin{table}[ht]
\caption{Accepted \texttt{agent\_version} after each coding round. A repeated version means the artifact did not change. On \texttt{no\_evolve} that is every round.}
\label{tab:codingver}
\begin{center}
\scriptsize
\setlength{\tabcolsep}{3.5pt}
\begin{tabular}{llrllll}
\toprule
Model & Method & Seed & r0 & r1 & r2 & r3 \\
\midrule
DeepSeek & no\_evolve & 42 & $M_0$ & $M_0$ & $M_0$ & $M_0$ \\
DeepSeek & no\_evolve & 43 & $M_0$ & $M_0$ & $M_0$ & $M_0$ \\
DeepSeek & no\_evolve & 44 & $M_0$ & $M_0$ & $M_0$ & $M_0$ \\
DeepSeek & gepa & 42 & $M_0$ & $M_1$ & $M_1$ & $M_1$ \\
DeepSeek & gepa & 43 & $M_0$ & $M_1$ & $M_1$ & $M_1$ \\
DeepSeek & gepa & 44 & $M_0$ & $M_1$ & $M_2$ & $M_2$ \\
DeepSeek & ace & 42 & $M_0$ & $M_0$ & $M_1$ & $M_1$ \\
DeepSeek & ace & 43 & $M_0$ & $M_0$ & $M_1$ & $M_1$ \\
DeepSeek & ace & 44 & $M_0$ & $M_0$ & $M_0$ & $M_0$ \\
DeepSeek & expel & 42 & $M_0$ & $M_1$ & $M_1$ & $M_1$ \\
DeepSeek & expel & 43 & $M_0$ & $M_1$ & $M_1$ & $M_1$ \\
DeepSeek & expel & 44 & $M_0$ & $M_0$ & $M_1$ & $M_1$ \\
GLM & no\_evolve & 42 & $M_0$ & $M_0$ & $M_0$ & $M_0$ \\
GLM & no\_evolve & 43 & $M_0$ & $M_0$ & $M_0$ & $M_0$ \\
GLM & no\_evolve & 44 & $M_0$ & $M_0$ & $M_0$ & $M_0$ \\
GLM & gepa & 42 & $M_0$ & $M_1$ & $M_1$ & $M_2$ \\
GLM & gepa & 43 & $M_0$ & $M_1$ & $M_2$ & $M_2$ \\
GLM & gepa & 44 & $M_0$ & $M_1$ & $M_2$ & $M_3$ \\
GLM & ace & 42 & $M_0$ & $M_1$ & $M_1$ & --- \\
GLM & ace & 43 & $M_0$ & $M_1$ & $M_1$ & --- \\
GLM & ace & 44 & $M_0$ & $M_1$ & $M_1$ & $M_2$ \\
GLM & expel & 42 & $M_0$ & $M_1$ & $M_2$ & $M_3$ \\
GLM & expel & 43 & $M_0$ & $M_1$ & $M_2$ & $M_2$ \\
GLM & expel & 44 & $M_0$ & $M_0$ & $M_1$ & $M_2$ \\
MiniMax & no\_evolve & 42 & $M_0$ & $M_0$ & $M_0$ & $M_0$ \\
MiniMax & no\_evolve & 43 & $M_0$ & $M_0$ & $M_0$ & $M_0$ \\
MiniMax & no\_evolve & 44 & $M_0$ & $M_0$ & $M_0$ & $M_0$ \\
MiniMax & gepa & 42 & $M_0$ & $M_0$ & $M_0$ & $M_0$ \\
MiniMax & gepa & 43 & $M_0$ & $M_1$ & $M_2$ & $M_3$ \\
MiniMax & gepa & 44 & $M_0$ & $M_1$ & $M_1$ & $M_1$ \\
MiniMax & ace & 42 & $M_0$ & $M_1$ & $M_2$ & $M_3$ \\
MiniMax & ace & 43 & $M_0$ & $M_1$ & $M_2$ & $M_3$ \\
MiniMax & ace & 44 & $M_0$ & $M_1$ & $M_2$ & $M_3$ \\
MiniMax & expel & 42 & $M_0$ & $M_1$ & $M_1$ & $M_2$ \\
MiniMax & expel & 43 & $M_0$ & $M_1$ & $M_1$ & $M_1$ \\
MiniMax & expel & 44 & $M_0$ & $M_1$ & $M_2$ & $M_3$ \\
\bottomrule
\end{tabular}
\end{center}
\end{table}

\begin{table}[ht]
\caption{Consecutive coding held-out passes with the same \texttt{agent\_version} and an absolute gap of at least 10 pp. The accepted artifact did not change between the two scores.}
\label{tab:codingjump}
\begin{center}
\scriptsize
\setlength{\tabcolsep}{3pt}
\begin{tabular}{llrllrrr}
\toprule
Model & Method & Seed & Version & Rounds & From & To & Gap \\
\midrule
DeepSeek & no\_evolve & 44 & $M_0$ & 2--3 & 65.0 & 52.5 & -12.5 \\
DeepSeek & gepa & 44 & $M_2$ & 2--3 & 67.5 & 42.5 & -25.0 \\
DeepSeek & ace & 42 & $M_0$ & 0--1 & 67.5 & 52.5 & -15.0 \\
GLM & no\_evolve & 42 & $M_0$ & 1--2 & 72.5 & 60.0 & -12.5 \\
GLM & no\_evolve & 42 & $M_0$ & 2--3 & 60.0 & 70.0 & +10.0 \\
GLM & no\_evolve & 44 & $M_0$ & 2--3 & 75.0 & 60.0 & -15.0 \\
GLM & ace & 44 & $M_1$ & 1--2 & 70.0 & 82.5 & +12.5 \\
GLM & expel & 44 & $M_0$ & 0--1 & 60.0 & 82.5 & +22.5 \\
MiniMax & no\_evolve & 44 & $M_0$ & 1--2 & 37.5 & 27.5 & -10.0 \\
MiniMax & no\_evolve & 44 & $M_0$ & 2--3 & 27.5 & 40.0 & +12.5 \\
MiniMax & gepa & 42 & $M_0$ & 1--2 & 30.0 & 42.5 & +12.5 \\
MiniMax & gepa & 42 & $M_0$ & 2--3 & 42.5 & 30.0 & -12.5 \\
MiniMax & expel & 42 & $M_1$ & 1--2 & 35.0 & 45.0 & +10.0 \\
MiniMax & expel & 43 & $M_1$ & 2--3 & 55.0 & 42.5 & -12.5 \\
\bottomrule
\end{tabular}
\end{center}
\end{table}

\begin{table}[ht]
\caption{DeepSeek coding failure counts by round. The count is \texttt{max\_steps}. \texttt{verifier\_fail} is absent on every pass.}
\label{tab:codingfail}
\begin{center}
\scriptsize
\setlength{\tabcolsep}{4pt}
\begin{tabular}{llrrrr}
\toprule
Method & Seed & r0 & r1 & r2 & r3 \\
\midrule
no\_evolve & 42 & 13 & 13 & 14 & 16 \\
no\_evolve & 43 & 15 & 14 & 15 & 16 \\
no\_evolve & 44 & 15 & 15 & 14 & 19 \\
gepa & 42 & 15 & 9 & 10 & 12 \\
gepa & 43 & 15 & 10 & 11 & 10 \\
gepa & 44 & 14 & 10 & 13 & 23 \\
ace & 42 & 13 & 19 & 9 & 8 \\
ace & 43 & 14 & 11 & 12 & 11 \\
ace & 44 & 16 & 14 & 17 & 16 \\
expel & 42 & 12 & 10 & 8 & 9 \\
expel & 43 & 14 & 9 & 9 & 12 \\
expel & 44 & 14 & 13 & 11 & 10 \\
\bottomrule
\end{tabular}
\end{center}
\end{table}
\section{Worked examples}
\label{app:examples}

The three task cards are evolution-split tasks. They are not held-out items, and they omit gold files, expected SQL, and unit-test oracles. The last card is one stored DeepSeek \texttt{no\_evolve} held-out trace pair; its instruction and written files are not reproduced.

\newcommand{\excard}[2]{%
\begin{tcolorbox}[colback=white,colframe=black!55,boxrule=0.5pt,arc=1pt,left=4pt,right=4pt,top=3pt,bottom=3pt,fonttitle=\small\bfseries,title={#1}]
\small
\begin{tabularx}{\linewidth}{@{}>{\raggedright\arraybackslash}p{0.22\linewidth}X@{}}
#2
\end{tabularx}
\end{tcolorbox}}

\subsection{One task from each current suite}

\excard{Office-v1 evolution task}{%
Task & \texttt{ofc\_evo\_discover\_open\_tasks} \\
Family & \texttt{catalog\_discover}; verification class \texttt{contract} \\
Instruction & ``List iteration-12 files and publish catalog\_index.md.'' \\
Observation & Two catalog files: \texttt{catalog/iter12.md}, \texttt{catalog/owners.md}. The gold \texttt{output/catalog\_index.md} is not in the observation. \\
Tools & \texttt{list\_files}, \texttt{read\_file}, \texttt{write\_file}, \texttt{publish\_artifact}, \texttt{grant\_access}, \texttt{submit} \\
Step cap & 20 \\
Success & Every terminal check passes. A failed \texttt{submit} returns the failed check names and leaves the episode open. \\
}

\excard{Coding-v1 evolution task}{%
Task & \texttt{coding\_fix\_hide\_loading\_gate} \\
Family & \texttt{fix\_bug}; verification class \texttt{executable} \\
Instruction & Chinese in the pack; English paraphrase: change \texttt{show\_realname\_gate} so a query that is still loading does not open the gate; only \texttt{required} or \texttt{unavailable} opens it. \\
Workspace & \texttt{src/solution.py} \\
Step cap & 12 \\
Success & Unit tests in the closed workspace pass. The gold patch is not shown. \\
}

\excard{Manufacturing-v1 evolution task}{%
Task & \texttt{manufacturing\_x\_tax\_2025\_personal} \\
Family & \texttt{tax} \\
Question & Chinese in the pack; English paraphrase: the group's 2025 personal-income-tax total. \\
Data & Real-derived tables; counterparty, company, and department names pseudonymized. \\
Step cap & 12 on this repeat \\
Success & Deterministic score at least 80. A miss is \texttt{below\_threshold} or \texttt{max\_steps}; \texttt{verifier\_fail} is not a manufacturing failure. The expected number and SQL are not shown. \\
}

\subsection{What \texttt{evolve()} actually returns}

\excard{Frozen control: DeepSeek \texttt{no\_evolve}, office seed 42}{%
Update & None. \texttt{evolve()} returns the same agent. \\
Stored files & An evolve input for rounds 1--3; no evolve-artifact file. \\
Version & $M_0$ after every round. \\
}

\excard{0 pp accept: DeepSeek \texttt{ace\_llm}, office seed 42, round 3}{%
Gate & Validation $0.350 \to 0.350$ under \texttt{min\_delta}${=}0$; accept (Table~\ref{tab:v1gate}). \\
Version & $M_2 \to M_3$ \\
Carrier & Text playbook. Round 3 keeps the four round-2 bullets and adds a fifth. \\
Added bullet & Before the final \texttt{publish\_artifact} that precedes submission, cross-check the draft against every task requirement and output rule, and edit the file to prevent \texttt{submit\_constraints\_failed}. \\
}

\subsection{One held-out task, same $M_0$, two scores}

\excard{Rescore: DeepSeek \texttt{no\_evolve}, office seed 42}{%
Task & \texttt{ofc\_hold\_discover\_rfq\_archive} (\texttt{catalog\_discover}) \\
Round 2 & $M_0$, 20 steps, fail (\texttt{max\_steps}) \\
Round 3 & $M_0$, 20 steps, pass \\
Effect & One task is 2.5 pp. This pair is part of the seed-42 path from 27.5\% at round 2 to 45\% at round 3. \\
}
\section{Archived ticket-gym mock}
\label{sec:office}

This section is not \texttt{office-v1}. The hosted matrix in Section~\ref{sec:officev1} uses catalog tools (\texttt{list\_files}, \texttt{read\_file}, \texttt{write\_file}, \texttt{publish\_artifact}, \texttt{grant\_access}, \texttt{submit}) and held-out $n{=}40$. The numbers below are the earlier in-process ticket mock (held-out $n{=}20$). They are not ranked against Table~\ref{tab:v1grid}.

Ticket-gym is a first-party in-process tool world: \texttt{search}, \texttt{read}, \texttt{write\_file}, \texttt{update\_ticket}, \texttt{auth\_refresh}, \texttt{list\_tickets}, \texttt{grant\_access}, \texttt{submit}. The verifier is deterministic (files, ticket state, submit args), and the same \texttt{task\_id} never crosses evolution / validation / held-out. Figure~\ref{fig:pack} visualizes the split; Table~\ref{tab:pack} lists its counts together with the inbox mock and the \texttt{office\_work} expansion. Evolution and validation are in-distribution only; held-out adds six \texttt{access\_denied} items (\texttt{ood=true}). Dual is therefore learnable from self failures; the OOD recover tool cannot be practiced on evolution.

\begin{figure}[ht]
\centering
\includegraphics[width=0.75\linewidth]{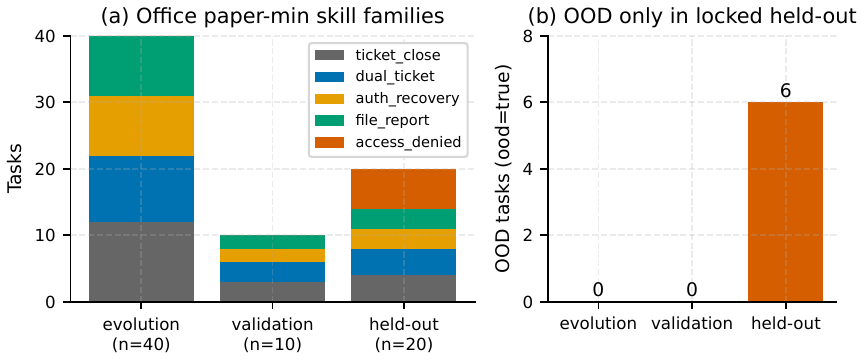}
\caption{Archived ticket-gym, not office-v1. Split is 40/10/20 and held-out $n{=}20$. OOD is locked in held-out.}
\label{fig:pack}
\end{figure}

\begin{table}[ht]
\caption{Archived ticket-gym family counts, held-out $n{=}20$, not office-v1. OOD families are held-out only. The same table lists the inbox mock and the \texttt{office\_work} expansion.}
\label{tab:pack}
\begin{center}
\scriptsize
\begin{tabular}{llccc}
\toprule
Env & Family & Evo & Val & Held-out \\
\midrule
ticket-gym & \texttt{ticket\_close} & 12 & 3 & 4 \\
ticket-gym & \texttt{dual\_ticket} & 10 & 3 & 4 \\
ticket-gym & \texttt{auth\_recovery} & 9 & 2 & 3 \\
ticket-gym & \texttt{file\_report} & 9 & 2 & 3 \\
ticket-gym & \texttt{access\_denied} (OOD) & 0 & 0 & 6 \\
\midrule
inbox & \texttt{triage} & 16 & 4 & 8 \\
inbox & \texttt{reply\_needed} & 12 & 3 & 3 \\
inbox & \texttt{thread\_pair} & 12 & 3 & 3 \\
inbox & \texttt{quarantine} (OOD) & 0 & 0 & 6 \\
\midrule
\texttt{office\_work} & \texttt{ticket\_close} & 6 & 2 & 4 \\
\texttt{office\_work} & \texttt{issue\_triage} & 8 & 2 & 3 \\
\texttt{office\_work} & \texttt{release\_notes} & 7 & 2 & 2 \\
\texttt{office\_work} & \texttt{postmortem} & 7 & 2 & 2 \\
\texttt{office\_work} & \texttt{diff\_review} & 6 & 1 & 2 \\
\texttt{office\_work} & \texttt{table\_brief} & 6 & 1 & 2 \\
\texttt{office\_work} & \texttt{access\_denied} (OOD) & 0 & 0 & 3 \\
\texttt{office\_work} & \texttt{checklist\_bound} (OOD) & 0 & 0 & 2 \\
\bottomrule
\end{tabular}
\end{center}
\end{table}

\paragraph{In-distribution families.}
\texttt{ticket\_close}: find a policy, write a summary, close one ticket. \texttt{dual\_ticket}: write a bundle covering two topics and close \emph{two} tickets with distinct notes; an incomplete \texttt{submit} does not end the episode. \texttt{auth\_recovery}: auth can expire again after a successful refresh. \texttt{file\_report}: report path plus required substrings.

\paragraph{OOD family.}
\texttt{access\_denied} ($n{=}6$ on held-out only): a tool returns \texttt{permission\_denied} until \texttt{grant\_access}, which is \emph{not} \texttt{auth\_refresh}. Evolution never contains this family. Transfer here is ``new recover tool,'' not a new domain.

Canonical heuristic $M_0$ is ticket-close only, including \texttt{auth\_refresh} on expiry. It does not know dual-ticket close or \texttt{grant\_access}. On this pack that floor is 50\% (10/20): the four ticket-close, three auth, and three file items, and none of dual or OOD.

\subsection{Why the protocol constraints}
\label{sec:rationale}

Each check maps a provenance or evaluation failure to a stored eligibility flag.

A teacher dump on the same tasks can look like a large Gain with no successful self rollout. The NL2SQL paper-min pack is the control: teacher \texttt{schema\_memory} reaches 100\%, self traces stay at 0\% (\texttt{anchor\_floor}; Appendix~\ref{app:worlds}). Teacher and self traces therefore occupy separate labeled boards. If \texttt{evolve()} sees held-out ids or gold fields, Success is leakage: the runner hashes the sorted held-out \texttt{task\_id} list and scans dumps. Training on held-out is not a TEB measurement.

Scoring a live server whose \texttt{artifact\_id} differs from the value returned by \texttt{evolve()} breaks artifact binding. The discarded DGM-inspired run listed \texttt{grant\_access\_on\_denied} in the mutation ISA and reached 80\% / OOD $6/6$ without OOD traces, so it is excluded (Appendix~\ref{app:native}). A B2 fullft and a B0 tip are different objects; the same evolver on a heuristic actor, frozen-3B wrapper, and native runtime is three cards.

Without rollback, wrapper GEPA~\citep{agrawal2025gepa} / Letta~\citep{packer2023memgpt} / DGM~\citep{zhang2025dgm} would look evolved; they reject every round, and \texttt{dgm\_native} accepts with validation 70\% while held-out stays 50\% (Appendix~\ref{app:gates}). Turning the gate off on B0 does not move those cards. $R{=}1$ cannot separate a jump from a loop; on this pack the heuristic jump is at r1, so $R{=}3$ is the eligibility floor. $M_0$ outside $[0.10,0.80]$ cannot show Gain and is flagged.

The default mix is \texttt{all} (successes and failures) because failure traces carry information that success traces alone do not encode---on ticket-gym, dual-ticket learning relies on recoverable failed \texttt{submit}s that expose the missing second ticket. Historical weight \texttt{success\_only} artifacts are excluded with their parent card because their trace actor and evaluated actor differ; they cannot support a mix claim.

The default curriculum is \texttt{none}~\citep{bengio2009curriculum} (static chunks) because static ordering keeps rounds comparable: every seed sees the same evolution ids in the same round, so round-level differences are attributable to the evolver, not to task selection. The \texttt{failure\_family} ablation reorders remaining evolution ids toward families that failed on prior traces; it does not raise heuristic held-out because dual is already saturated and OOD never enters evolution. It is a labeled ablation, not a candidate default.

The default schedule is \texttt{accumulate\_chunks}: each evolution id is collected in at most one round, and \texttt{evolve()} sees traces from rounds $1{\ldots}r$. \texttt{replay} reuses the first 14 evolution ids across all rounds instead; it is marked ineligible (\texttt{round\_schedule}) because a second pass over the same tasks cannot introduce new failure modes. \texttt{accumulate\_chunks} ensures that each round contributes new traces to the evolving payload, so a system that learns from novel failures is not starved of them.

The archived ticket-gym uses a 40/10/20 paper-min floor. Office-v1 and manufacturing-v1 are 80/20/40. That floor ($R{=}3$, $\ge 3$ seeds) is a statistical minimum, not an optimum. Held-out $n{=}20$ gives each family at least 3--4 items so a family rate is not a single-task coin flip; evolution $n{=}40$ gives $R{=}3$ at least 13 new ids per round under \texttt{accumulate\_chunks}, so a round is not a single-batch update. Fewer than these and a card's Success~\% and Gain are noisy enough to obscure the family split that the protocol exists to expose.

TEB does not require weight updates or emit a scalar ``evolution IQ.'' Seeds do not alter held-out membership, failed self rollouts are not replaced with teacher traces, held-out inference is outside the update \texttt{budget\_class}, and main-table verification is deterministic rather than LLM-as-judge~\citep{zheng2023judge}.

\subsection{Success, Gain, and actors}

Success $\%= 100 \times \#\{\text{held-out pass}\}/\#\{\text{held-out}\}$. Gain (pp) $= 100 \times (\mathrm{Succ}(M_r)-\mathrm{Succ}(M_0))$, with $M_0$ \emph{this-run} unless the card is eval-only with a frozen floor. Rank only inside $(\texttt{env}, \texttt{budget\_class}, \texttt{trace\_origin}, \texttt{actor})$. An adapter name is a retrofit; the actor is who emits the tool call.

\subsection{Setup of the archived mock}
\label{sec:setup}

All ticket-gym numbers use tasks 40/10/20, $R{=}3$, \texttt{trace\_origin=self}, mix \texttt{all}, schedule \texttt{accumulate\_chunks}, gate \texttt{min\_delta} ${=}0$, unless labeled otherwise. Evolution collection is chunked at 14 tasks per round. Episode cap is 20 env steps. Seeds $\{42,43,44\}$ shuffle evolution order, not held-out membership.

\textbf{Heuristic B0.} Actor: \texttt{SandboxAgent}. Evolvers: \texttt{no\_evolve}, \texttt{prompt\_memory}, \texttt{success\_memory}, stand-in \texttt{reflexion}~\citep{shinn2023reflexion} / \texttt{expel}~\citep{zhao2024expel} / \texttt{awm}~\citep{wang2024awm}. No train GPU. Stand-in Reflexion maps failure traces onto executable tips (e.g.\ \texttt{workflow:dual\_ticket\_close}); there is no LLM judge.

\textbf{Weight B2.} Actor: Qwen2.5-3B-Instruct~\citep{qwen25} after self-trace full fine-tune. Eval binds \texttt{evolve}'s artifact. Gain vs.\ frozen heuristic $M_0{=}0.50$ (eval-only of a trained checkpoint is not a re-measure of untrained 3B $M_0$).

\textbf{Verbal B0 wrappers.} Same frozen 3B, TEB-stuffed tips, seed 42. Gain vs.\ \emph{this-run} $M_0$ (measured $0.45 / 0.50 / 0.55$ across cards; not the heuristic $0.50$). Wrapper $M_0$ is unbound relative to the heuristic floor, so these rows are not ranked into the B0 heuristic card. These are not native runs of~\citet{packer2023memgpt,agrawal2025gepa,zhang2025dgm,zhai2025agentevolver}.

\textbf{Native pairing} (separate board): Letta~\citep{packer2023memgpt} messages API as actor; DGM~\citep{zhang2025dgm} genome compiled onto \texttt{SandboxAgent} without leaking OOD tool names in the mutation ISA; the same Reflexion-style~\citep{shinn2023reflexion} wrapper loop on a larger chat model (\texttt{qwen-plus}).

\texttt{inbox} is a second first-party mock with the same 40/10/20 shape (Section~\ref{sec:inbox}). The NL2SQL paper-min pack is a teacher-contrast negative control, not a third headline world (Appendix~\ref{app:worlds}).

\subsection{Summary boards and expansion families}
\label{sec:results}

Table~\ref{tab:boards} separates the eligible heuristic ticket-gym card from one-seed wrapper contrasts and the incomplete weight attempt. Rows are boards, not methods to sort. The expansion pack is not another board: it has $M_0{=}0.20$ and different families, so 75\% is not a denser 70\%. Table~\ref{tab:workfam} is that pack's family counts. Table~\ref{tab:officework} is the same pack's Success summary.

\begin{table}[ht]
\caption{Archived ticket-gym mock, held-out $n{=}20$. Not an office-v1 score. Heuristic rows are three seeds. Wrapper rows are one seed. The weight attempt submitted no training.}
\label{tab:boards}
\begin{center}
\scriptsize
\begin{tabular}{lcccccc}
\toprule
Board & Succ. & Gain & dual & file & OOD & Seeds \\
\midrule
B0 reflexion (heuristic) & 70\% & $+20$ vs $0.50$ & $4/4$ & $3/3$ & $0/6$ & 3 \\
B0 other (heuristic) & 50\% & $+0$ & $0/4$ & $3/3$ & $0/6$ & 3 \\
Weight attempt (ineligible) & 50\% & $+0$ vs $0.50$ & $0/4$ & $3/3$ & $0/6$ & 1 \\
3B \texttt{reflexion\_llm} & 55\% & $+10$ vs $0.45$ & $0/4$ & $0/3$ & $4/6$ & 1 \\
3B other wrappers & $45$--$55\%$ & $+0$ this-run & $0/4$ & $0/3$ & $3$--$4/6$ & 1 \\
\bottomrule
\end{tabular}
\end{center}
\end{table}

\begin{table}[ht]
\caption{Archived ticket-gym expansion pack \texttt{office\_work}, not office-v1. Held-out $n{=}20$, three identical seeds. Heuristic B0 only. Not ranked vs.\ Table~\ref{tab:boards}.}
\label{tab:workfam}
\begin{center}
\scriptsize
\begin{tabular}{lccc}
\toprule
Family & $n$ & $M_0$ / other B0 & reflexion \\
\midrule
\texttt{ticket\_close} & 4 & $4/4$ & $4/4$ \\
\texttt{issue\_triage} & 3 & $0/3$ & $3/3$ \\
\texttt{release\_notes} & 2 & $0/2$ & $2/2$ \\
\texttt{postmortem} & 2 & $0/2$ & $2/2$ \\
\texttt{diff\_review} & 2 & $0/2$ & $2/2$ \\
\texttt{table\_brief} & 2 & $0/2$ & $2/2$ \\
\texttt{access\_denied} (OOD) & 3 & $0/3$ & $0/3$ \\
\texttt{checklist\_bound} (OOD) & 2 & $0/2$ & $0/2$ \\
\midrule
Held-out Success & 20 & 20\% & 75\% ($+55$pp) \\
\bottomrule
\end{tabular}
\end{center}
\end{table}

\subsection{Round curves}

Heuristic stand-in Reflexion jumps at r1 and saturates (70\%). Other B0 methods stay at the $M_0$ floor. Among one-seed wrappers, only \texttt{reflexion\_llm} moves ($45{\to}55$); GEPA/Letta/DGM wrappers roll back to this-run $M_0$; \texttt{agent\_evolver} $45{\to}55$ occurs under rollback and is $M_0$ jitter, not a second $+10$pp. Wrapper \texttt{file\_report} stays $0/3$.

\subsection{Mechanism: what heuristic failures teach}

At heuristic $M_0$, ticket-close, auth, and file are already solved. Stand-in Reflexion's $+20$pp is dual-ticket going $0/4{\to}4/4$ at r1 from \emph{self} incomplete-submit traces; OOD stays $0/6$ because that family is absent from evolution.

Figure~\ref{fig:traces} is a live \texttt{SandboxAgent} rollout on two locked held-out ids, not a 3B completion log. On dual (\texttt{ofc\_hold\_du\_001}), $M_0$ searches once then repeatedly \texttt{submit}s a single ticket; the verifier returns \texttt{submit\_constraints\_failed} and the episode continues to \texttt{max\_steps}. After the stand-in writes \texttt{workflow:dual\_ticket\_close}, the same id is solved by two \texttt{update\_ticket}s with distinct notes. On OOD (\texttt{ofc\_hold\_ad\_001}), $M_0$ never calls \texttt{grant\_access} and dies in \texttt{permission\_denied}; the recover tool is a one-step unlock, after which the ticket-close loop succeeds.

\begin{figure}[ht]
\centering
\includegraphics[width=0.75\linewidth]{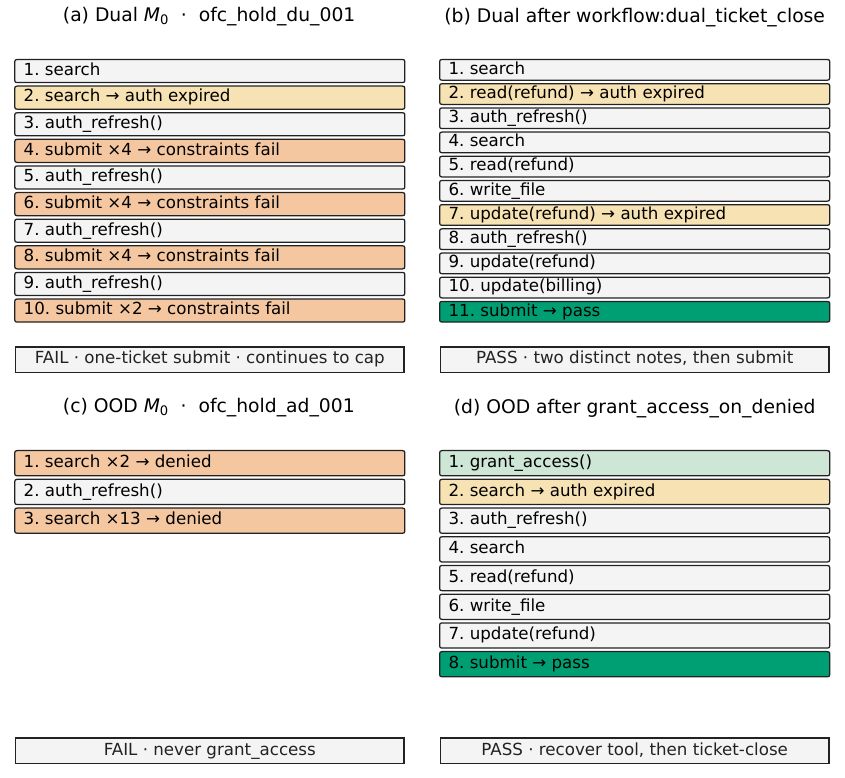}
\caption{Archived ticket-gym tool calls, not office-v1. Locked held-out ids (\texttt{SandboxAgent}, not a 3B log). (a)~$M_0$ dual: recoverable failed \texttt{submit}, second ticket never closed. (b)~Same id after \texttt{workflow:dual\_ticket\_close}. (c)~$M_0$ OOD: no \texttt{grant\_access}. (d)~OOD after \texttt{grant\_access\_on\_denied}. Consecutive identical calls are collapsed ($\times n$).}
\label{fig:traces}
\end{figure}

A hygiene run that drops \texttt{grant\_access} from the serving prompt puts no-hint 3B $M_0$ at 30\%, OOD $0/6$ (Table~\ref{tab:hygiene}). The hinted wrapper's $M_0$ OOD came from the prompt, not a learned recover skill. Frozen wrapper cards keep the hinted prompt so they reproduce; the no-hint $M_0$ is reported separately.

\subsection{Why $R{=}3$ is enough on ticket-gym}

Heuristic Reflexion jumps at r1 and saturates (70\% / dual $4/4$). Extra dual practice (\texttt{failure\_family}) and \texttt{replay} of the first 14 evolution ids do not raise held-out, because dual is already solved for the stand-in and OOD \texttt{access\_denied} is still absent from evolution. We therefore keep $R{=}3$ as the replication floor rather than grinding more rounds.

\subsection{Inbox mock}
\label{sec:inbox}

To check that the protocol is not specific to ticket-gym, \texttt{inbox} provides a second in-process mock (list, read, label, archive, reply, quarantine) with the same 40/10/20 paper-min shape, $R{=}3$, and three seeds (Table~\ref{tab:inbox}). Canonical $M_0$ is triage-only (40\%). B0 stand-in Reflexion reaches 70\% / $+30$pp by unlocking reply $3/3$ and thread $3/3$, and never OOD quarantine ($0/6$). Other B0 methods stay at 40\%.

As in ticket-gym, the stand-in learns an in-distribution family exposed by failed submits. Historical inbox weight grids are excluded because first-round traces came from \texttt{InboxAgent}, not the evaluated 3B. Inbox and ticket-gym remain separate boards; inbox is not a production mail client.

\subsection{Seed, wrapper, inbox, and hygiene tables}
\label{app:cards}

These tables record the archived heuristic seeds, the seed-42 3B wrappers, the inbox board, and the prompt-hygiene check. They are not a ranking and not extra models. Step-token columns in the runner are \texttt{estimate}s (Limitations); we do not treat them as a serving benchmark. Wall-clock is process time of the seed-42 3B loops only. Wrapper round paths below are held-out Success~\% of the \emph{accepted} agent (rollback keeps $M_{r-1}$; re-eval jitter is labeled). Native pairing is Table~\ref{tab:native}.

\begin{table}[ht]
\caption{Archived ticket-gym heuristic B0, not office-v1. Held-out Success~\% by seed. Other B0 methods are \texttt{no\_evolve}, \texttt{prompt\_memory}, \texttt{success\_memory}, \texttt{expel}, \texttt{awm}.}
\label{tab:b0seeds}
\begin{center}
\scriptsize
\begin{tabular}{lccccc}
\toprule
Evolver & s42 & s43 & s44 & dual & OOD \\
\midrule
\texttt{reflexion} (stand-in) & 70 & 70 & 70 & $4/4$ & $0/6$ \\
\texttt{no\_evolve} & 50 & 50 & 50 & $0/4$ & $0/6$ \\
\texttt{prompt\_memory} & 50 & 50 & 50 & $0/4$ & $0/6$ \\
\texttt{success\_memory} & 50 & 50 & 50 & $0/4$ & $0/6$ \\
\texttt{expel} & 50 & 50 & 50 & $0/4$ & $0/6$ \\
\texttt{awm} & 50 & 50 & 50 & $0/4$ & $0/6$ \\
\bottomrule
\end{tabular}
\end{center}
\end{table}

\begin{table}[ht]
\caption{Archived ticket-gym 3B tip-wrappers, not office-v1. Seed 42. Columns r0--r3 are held-out Success~\% of the accepted agent. \texttt{reflexion\_llm} r3 rolls back (card is M2 55\%). \texttt{agent\_evolver} $45{\to}50{\to}55$ occurs under three rollbacks (jitter), not a second $+10$pp. Dual $0/4$ and file $0/3$ on every row. Wall is loop process time, not time-to-first-token.}
\label{tab:wrappers}
\begin{center}
\scriptsize
\begin{tabular}{lccccccc}
\toprule
Loop & r0 & r1 & r2 & r3 & Gain & OOD & Wall (h) \\
\midrule
\texttt{reflexion\_llm} & 45 & 55 & 55 & 55 & $+10$ & $4/6$ & 4.0 \\
\texttt{gepa\_llm} & 50 & 50 & 50 & 50 & $+0$ & $3/6$ & 3.8 \\
\texttt{letta} wrapper & 55 & 55 & 55 & 55 & $+0$ & $4/6$ & 5.9 \\
\texttt{dgm} wrapper & 45 & 45 & 45 & 45 & $+0$ & $3/6$ & 5.7 \\
\texttt{agent\_evolver} & 45 & 50 & 55 & 55 & jitter & $4/6$ & 5.4 \\
\bottomrule
\end{tabular}
\end{center}
\end{table}

\begin{table}[ht]
\caption{Archived inbox mock, not office-v1. Held-out $n{=}20$. Not ranked vs.\ Table~\ref{tab:boards}. Reply $n{=}3$, thread $n{=}3$, OOD quarantine $n{=}6$.}
\label{tab:inbox}
\begin{center}
\scriptsize
\begin{tabular}{lcccccc}
\toprule
Board & Succ. & Gain & reply & thread & OOD & Seeds \\
\midrule
B0 reflexion & 70\% & $+30$ vs $0.40$ & $3/3$ & $3/3$ & $0/6$ & 3 \\
B0 other & 40\% & $+0$ & $0/3$ & $0/3$ & $0/6$ & 3 \\
\bottomrule
\end{tabular}
\end{center}
\end{table}

\begin{table}[ht]
\caption{Archived ticket-gym hygiene, not office-v1. Same frozen 3B, serving prompt with vs.\ without \texttt{grant\_access} hint. Eval-only ($R{=}0$). Dual $0/4$ on both rows. Table~\ref{tab:wrappers} keeps the hinted prompt.}
\label{tab:hygiene}
\begin{center}
\scriptsize
\begin{tabular}{lccccc}
\toprule
Prompt & Succ. & tc & auth & file & OOD \\
\midrule
Hinted wrapper $M_0$ & 45\% & $4/4$ & $2/3$ & $0/3$ & $3/6$ \\
No hint $M_0$ & 30\% & $4/4$ & $2/3$ & $0/3$ & $0/6$ \\
\bottomrule
\end{tabular}
\end{center}
\end{table}

\subsection{Weight-track provenance}
\label{app:weightpath}

Historical ticket-gym, inbox, and office-work weight grids are excluded. Their first
round collected ``self'' traces with local heuristic actors, then trained and
evaluated a 3B checkpoint; a scalar frozen heuristic $M_0$ did not bind that
change of actor. Artifact matching alone cannot repair the provenance mismatch.
The ticket-gym native attempt collects traces with the served 3B actor, but
its \texttt{prepare\_submit} round did not submit training and was rolled back;
the resulting 50\% / $+0$ card is an ineligible infrastructure outcome, not a
weight-method result. We therefore report no weight-track headline.

\subsection{Ticket-gym B0 ablations}
\label{app:ablate}

Gate-off, replay, and failure-family runs keep heuristic Reflexion at 70\% and
other B0 methods at 50\% (three seeds). Replay and failure-family remain labeled
contrasts. Historical weight-mix and hybrid artifacts inherit the trace-actor
provenance failure and are not ablation results.

\subsection{Ticket-gym gate decisions}
\label{app:gates}

Figure~\ref{fig:gates} is the accept/rollback heatmap. Table~\ref{tab:gates} adds the held-out consequence. Green in the figure is accept; orange is rollback. Office seed 42 unless labeled B0. Validation Success is the gate input; held-out Success is scored on the accepted agent after that decision.

\begin{table}[ht]
\caption{Archived ticket-gym gates, not office-v1 (Table~\ref{tab:v1gate}). A = accept, R = rollback. \texttt{dgm\_native} can accept a validation 70\% proposal while held-out stays 50\%. \texttt{qwen-plus} r3 accepts $0{\to}0$; the card is final 5\%, not the M2 peak.}
\label{tab:gates}
\begin{center}
\scriptsize
\begin{tabular}{lcccc}
\toprule
Card & r1 & r2 & r3 & Held-out note \\
\midrule
B0 reflexion & A & A & A & 70\% from r1 \\
\texttt{reflexion\_llm} & A & A & R & stays M2 55\% \\
\texttt{gepa}/\texttt{letta}/\texttt{dgm} wrap & R & R & R & this-run $M_0$ \\
\texttt{agent\_evolver} & R & R & R & $45{\to}55$ jitter \\
\texttt{letta\_native} & A & A & R & r1 is 10\%; final M2 jitter 65\% \\
\texttt{dgm\_native} & A & A & A & val 70\%, held-out 50\% \\
\texttt{qwen-plus} & A & A & A & r0--r3: $5,5,15,5$ \\
\bottomrule
\end{tabular}
\end{center}
\end{table}

\begin{figure}[ht]
\centering
\includegraphics[width=0.75\linewidth]{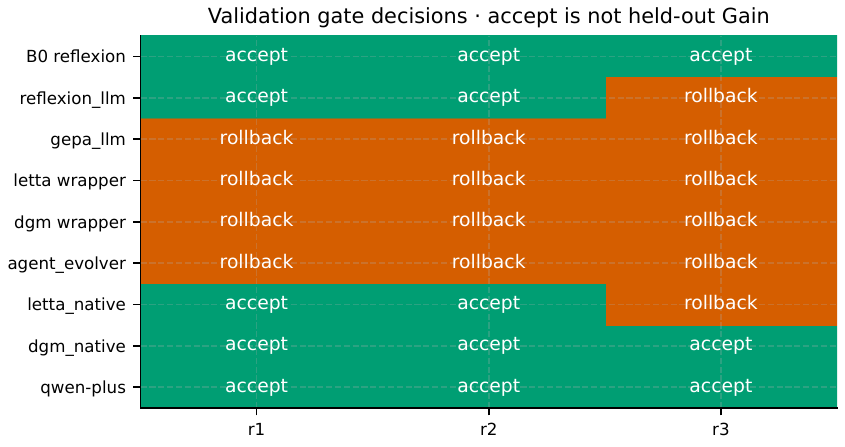}
\caption{Archived ticket-gym gates, not office-v1. Accept and rollback are not held-out Gain.}
\label{fig:gates}
\end{figure}

\subsection{Ticket-gym native pairing}
\label{app:native}

Not ranked vs.\ Table~\ref{tab:boards} (Table~\ref{tab:native}). \texttt{letta\_native}: this-run $M_0$ 60\%, final 65\% / $+5$pp, which is M2 re-eval jitter after r3 rollback (r1 held-out was 10\%); dual $2/4$ is the first 3B-backed dual on this pack, while \texttt{ticket\_close} stays $0/4$. \texttt{dgm\_native} (ISA does not name OOD tools): 50\% / $+0$pp. A discarded run that listed \texttt{grant\_access\_on\_denied} in the mutation ISA hit 80\% / OOD $6/6$ without OOD traces---wrong mechanism, not a result. Historical weight grids are audit artifacts, not results.

The scale probe in this freeze is the same Reflexion-style wrapper loop on \texttt{qwen-plus} (a larger chat model than the bound 3B; not a 7B/14B fullft). This-run $M_0$ is 5\%; peak M2 is 15\%; the card is final 5\% / $+0$pp / dual $0/4$ (r3 undid M2). A bigger decoder did not unlock dual the way the heuristic stand-in does. We do not headline the M2 peak.

\begin{table}[ht]
\caption{Archived ticket-gym native pairing, not office-v1. Seed 42. Columns r0--r3 are held-out Success~\% of the accepted agent. Dual/OOD are the frozen (final) card. Not ranked vs.\ Table~\ref{tab:boards}. \texttt{letta\_native} r3 rolls back: final 65\% is M2 re-eval jitter vs.\ r2 55\%. \texttt{qwen-plus} peak r2 15\% is not the card. \texttt{dgm\_native} actor is \texttt{SandboxAgent} (genome only). File: Letta/DGM $3/3$, qwen-plus $0/3$. Ticket-close: Letta $0/4$, DGM $4/4$, qwen-plus $0/4$.}
\label{tab:native}
\begin{center}
\scriptsize
\begin{tabular}{lccccccc}
\toprule
Card & r0 & r1 & r2 & r3 & Gain & dual & OOD \\
\midrule
\texttt{letta\_native} & 60 & 10 & 55 & 65 & $+5$ jitter & $2/4$ & $6/6$ \\
\texttt{dgm\_native} & 50 & 50 & 50 & 50 & $+0$ & $0/4$ & $0/6$ \\
\texttt{qwen-plus} & 5 & 5 & 15 & 5 & $+0$ & $0/4$ & $1/6$ \\
\bottomrule
\end{tabular}
\end{center}
\end{table}

\begin{figure}[ht]
\centering
\includegraphics[width=0.75\linewidth]{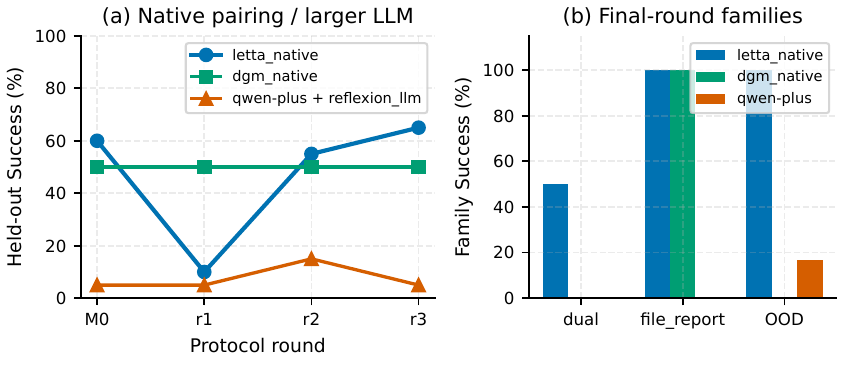}
\caption{Archived ticket-gym native pairing board, not office-v1. Not ranked vs.\ Table~\ref{tab:boards}.}
\label{fig:native}
\end{figure}

\subsection{NL2SQL teacher contrast}
\label{app:worlds}

NL2SQL paper-min 40/10/20~\citep{yu2018spider,li2023bird}, not the 12/3/9 snapshot. Table~\ref{tab:wenshu}: teacher \texttt{schema\_memory} $0{\to}100\%$ by r1 (three seeds), OOD $4/4$; self all 0\% (\texttt{anchor\_floor}). This is a labeled teacher contrast and a protocol floor, not an office-v1 result. Teacher 100\% is not mixed with any self board.

\begin{table}[ht]
\caption{NL2SQL teacher contrast, not office-v1. Held-out Success~\%, mean of seeds 42--44 (cells identical across seeds). Self has no successful SQL in traces.}
\label{tab:wenshu}
\begin{center}
\scriptsize
\begin{tabular}{lcc}
\toprule
Evolver & Teacher traces & Self traces \\
\midrule
\texttt{schema\_memory} & 100\% & 0\% \\
\texttt{prompt\_memory} & 0\% & 0\% \\
\texttt{no\_evolve} & 0\% & 0\% \\
\bottomrule
\end{tabular}
\end{center}
\end{table}

\begin{figure}[ht]
\centering
\includegraphics[width=0.75\linewidth]{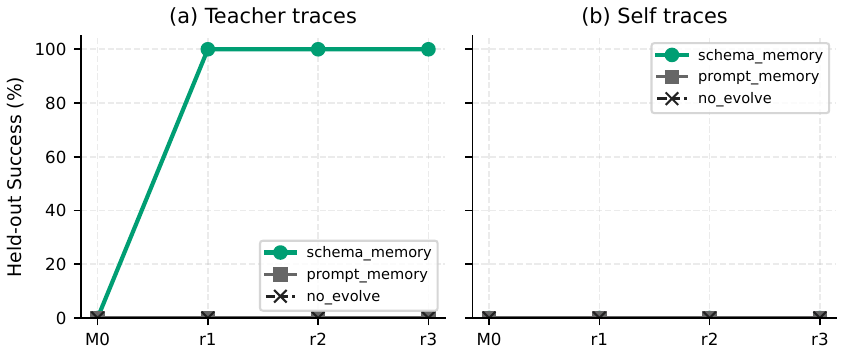}
\caption{NL2SQL teacher contrast, not office-v1. (a) Teacher traces. (b) Self traces, which contain no successful SQL, so every evolver stays at 0\%.}
\label{fig:wenshu}
\end{figure}

\subsection{Ticket-gym expansion pack}
\label{app:officework}

Same ticket-gym tools, new families mapped from public product-usage cases, 40/10/20. Family counts are Table~\ref{tab:workfam}. Env id \texttt{office\_work}, lane \texttt{contrast} (\texttt{pack\_not\_v1}). Not a fourth frozen actor board and not mixed with ticket-gym or inbox. Heuristic B0 (three seeds): measured $M_0$ 20\%; stand-in Reflexion 75\% / $+55$pp (in-dist $15/15$); other B0 stay 20\%; OOD \texttt{access\_denied} $0/3$ and \texttt{checklist\_bound} $0/2$ (Figure~\ref{fig:officework}, Table~\ref{tab:officework}). No valid weight result is reported.

\clearpage

\begin{table}[H]
\caption{Archived ticket-gym \texttt{office\_work} heuristic B0, not office-v1. Held-out Success~\%. Identical across seeds 42--44. Other B0: \texttt{no\_evolve}, \texttt{prompt\_memory}, \texttt{success\_memory}, \texttt{expel}, \texttt{awm}.}
\label{tab:officework}
\begin{center}
\scriptsize
\begin{tabular}{lcccc}
\toprule
Evolver & Success & Gain & in-dist & OOD \\
\midrule
\texttt{reflexion} (stand-in) & 75\% & $+55$pp & $15/15$ & $0/5$ \\
other B0 & 20\% & $+0$pp & $4/15$ (tc only) & $0/5$ \\
\bottomrule
\end{tabular}
\end{center}
\end{table}

\begin{figure}[H]
\centering
\includegraphics[width=0.85\linewidth]{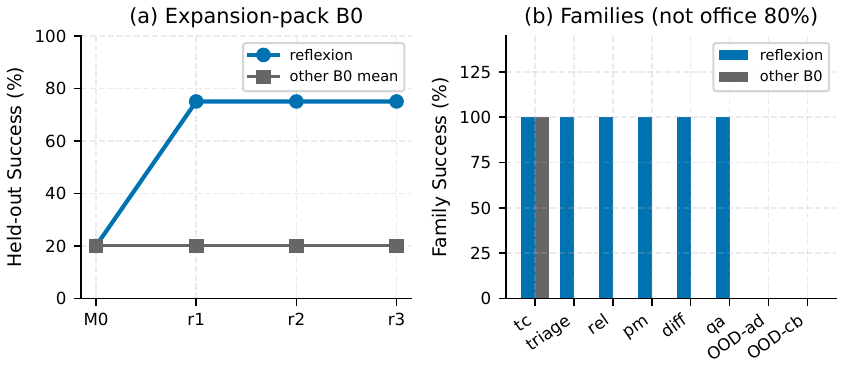}
\caption{Archived ticket-gym expansion pack, not office-v1. Not ranked vs.\ Table~\ref{tab:boards}.}
\label{fig:officework}
\end{figure}

\subsection{Cross-environment summary}
\label{app:summary}

Table~\ref{tab:summary} consolidates the frozen ticket-gym, inbox, NL2SQL, and \texttt{office\_work} cards. Rows are actor boards. Each row uses the 40/10/20 pack, $R{=}3$, the validation gate, and artifact binding. Seed counts are the last column: the 3B wrapper row is one seed, and the other rows are three. No single row is ranked above another across budget classes or environments.

\begin{table}[ht]
\caption{Archived cards across ticket-gym, inbox, NL2SQL, and \texttt{office\_work}. Not an office-v1 ranking. ``---'' = family not present in that environment.}
\label{tab:summary}
\begin{center}
\scriptsize
\setlength{\tabcolsep}{3pt}
\begin{tabular}{llcccccc}
\toprule
Env & Board & Budget & Succ. & Gain & In-dist family & OOD family & Seeds \\
\midrule
\multirow{2}{*}{ticket-gym} & B0 reflexion (heur.) & B0 & 70\% & $+20$ vs 0.50 & dual 4/4 & 0/6 & 3 \\
 & 3B reflexion\_llm wrap & B0 & 55\% & $+10$ vs 0.45 & dual 0/4 & 4/6 & 1 \\
\midrule
\multirow{2}{*}{inbox} & B0 reflexion (heur.) & B0 & 70\% & $+30$ vs 0.40 & reply 3/3, thread 3/3 & 0/6 & 3 \\
 & B0 other (heur.) & B0 & 40\% & $+0$ & 0/3 & 0/6 & 3 \\
\midrule
\multirow{2}{*}{NL2SQL} & Teacher schema\_mem & --- & 100\% & $+100$ vs 0 & OOD 4/4 & --- & 3 \\
 (contrast) & Self (all evolvers) & --- & 0\% & --- & 0/4 & --- & 3 \\
\midrule
office\_work & B0 reflexion (heur.) & B0 & 75\% & $+55$ vs 0.20 & in-dist 15/15 & 0/5 & 3 \\
\bottomrule
\end{tabular}
\end{center}
\end{table}

\end{document}